\documentclass[lettersize,journal]{IEEEtran}
\usepackage{amsmath,amsfonts}
\usepackage{algorithmic}
\usepackage{algorithm}
\usepackage{array}
\usepackage[caption=false,font=footnotesize,labelfont=rm,textfont=rm]{subfig}
\usepackage{textcomp}
\usepackage{stfloats}
\usepackage{url}
\usepackage{verbatim}
\usepackage{graphicx}
\usepackage{cite}
\usepackage{tikz}
\usepackage{pgfplots}
\usetikzlibrary{arrows.meta}
\usepackage{adjustbox}
\usepackage{hyperref}
\usepackage{subfig}
\usepackage{mathptmx}
\hypersetup{
    colorlinks=true,
    linkcolor=black,
    urlcolor=black,
}
\usetikzlibrary{positioning, calc, shadows}
\usepackage{xcolor}
\usepackage{tabularx}
\usepackage{makecell}

\usepackage{booktabs}
\usepackage{multirow}
\usepackage{siunitx}

\usepackage{color}

\pgfplotsset{compat=1.3}
\begin{document}

\title{An Empirical Analysis of Cooperative Perception \\ for Occlusion Risk Mitigation}

\author{Aihong Wang\textsuperscript{*}, Tenghui Xie\textsuperscript{*}, Fuxi Wen and Jun Li
\thanks{\textsuperscript{*}Aihong Wang and Tenghui Xie contributed equally to this work. The authors are with the School of Vehicle and Mobility,
Tsinghua University, China. \textit{(Corresponding author: Fuxi Wen)}}
\thanks{This work was partially supported by the National Key R\&D Program of China  under Grant 2024YFB2505803.}
}

\markboth{IEEE Internet of Things Journal,~Vol.~*, No.~*, \today}%
{Shell \MakeLowercase{\textit{et al.}}: A Sample Article Using IEEEtran.cls for IEEE Journals}


\maketitle

\begin{abstract}

Occlusions present a significant challenge for connected and automated vehicles, as they can obscure critical road users from perception systems. Traditional risk metrics often fail to capture the cumulative nature of these threats over time adequately. In this paper, we propose a novel and universal risk assessment metric, the Risk of Tracking Loss (RTL), which aggregates instantaneous risk intensity throughout occluded periods. This provides a holistic risk profile that encompasses both high-intensity, short-term threats and prolonged exposure. Utilizing diverse and high-fidelity real-world datasets, a large-scale statistical analysis is conducted to characterize occlusion risk and validate the effectiveness of the proposed metric. 
The metric is applied to evaluate different vehicle-to-everything (V2X) deployment strategies. Our study shows that full V2X penetration theoretically eliminates this risk, the reduction is highly nonlinear; a substantial statistical benefit requires a high penetration threshold of 75–90\%. To overcome this limitation, we propose a novel asymmetric communication framework that allows even non-connected vehicles to receive warnings. Experimental results demonstrate that this paradigm achieves better risk mitigation performance. We found that our approach at 25\% penetration outperforms the traditional symmetric model at 75\%, and benefits saturate at only 50\% penetration. This work provides a crucial risk assessment metric and a cost-effective, strategic roadmap for accelerating the safety benefits of V2X deployment.

\end{abstract}


 \section{Introduction}
\label{section:intro}

\begin{figure}[ht]
    \centering
    \subfloat[Static occlusion.]{
    \includegraphics[width=0.45\linewidth]{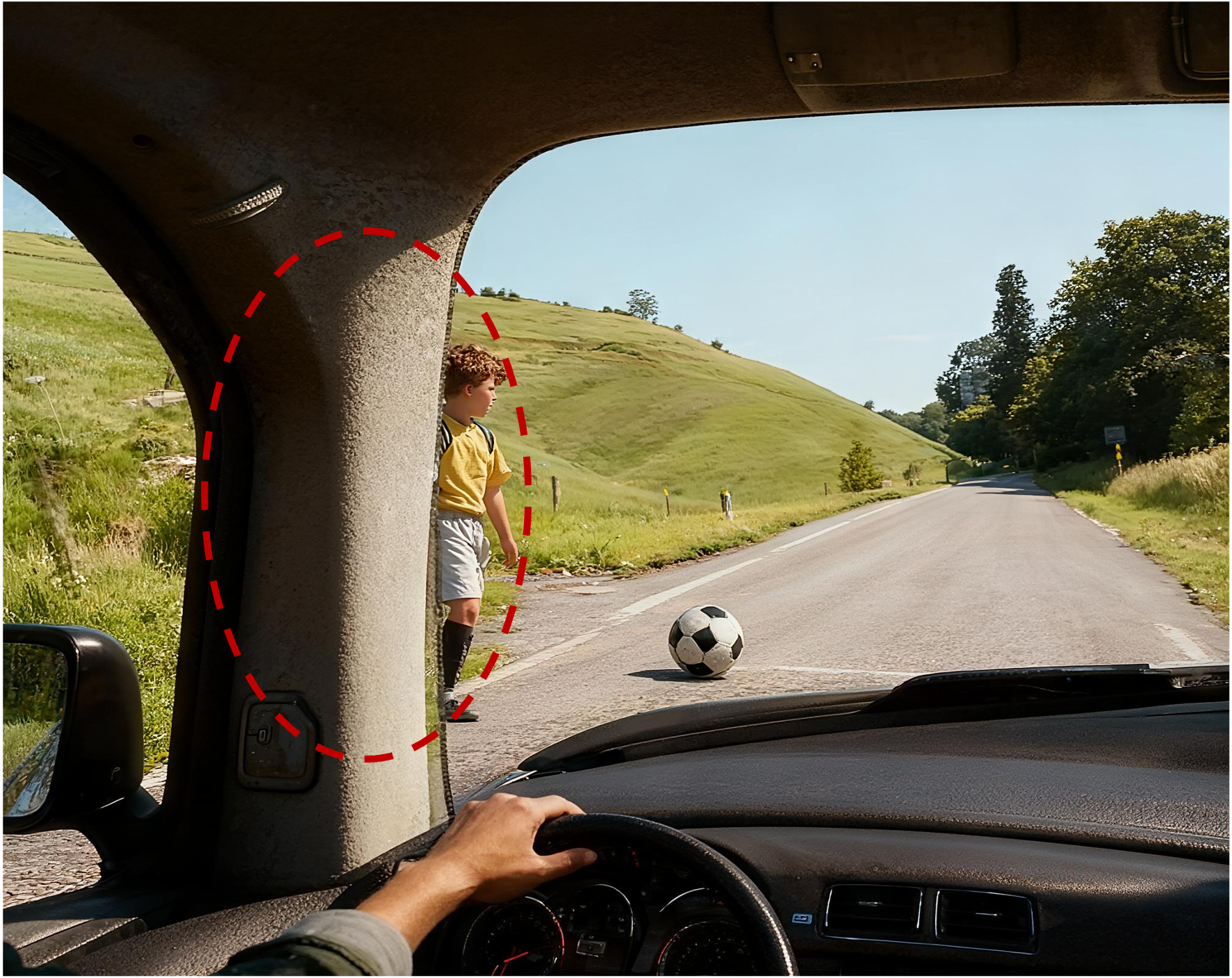}
    \label{fig:static_occlusion}}
    \subfloat[Dynamic occlusion.]{
    \includegraphics[width=0.45\linewidth]{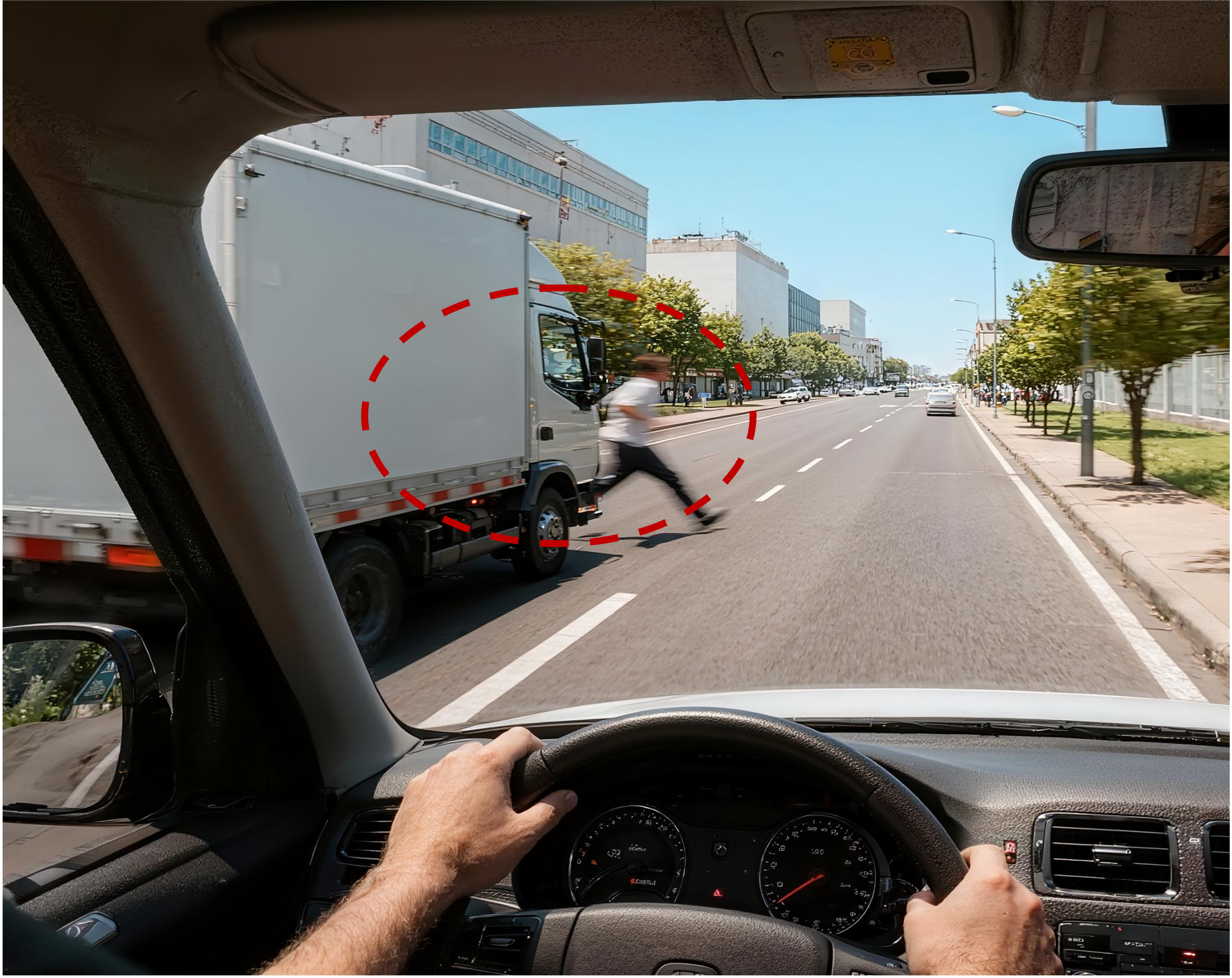}
    \label{fig:dynamic_occlusion}}
    \caption{Illustrative examples of critical sensory blind spots for a vehicle, with the resulting occluded regions circled in red. (a) A static occlusion caused by the A-pillar, which obscures a child about to retrieve a ball. (b) A dynamic occlusion caused by a large truck, which conceals a pedestrian running across the road from its front.}
    \label{fig:occlusions}
\end{figure}

Perceptual occlusion, a persistent safety challenge for human drivers, poses a more critical threat to autonomous systems, which lack the capacity for intuitive reasoning \cite{gilroy2019overcoming}. 
These sensory blind spots arise from two primary sources: static occlusions inherent to the vehicle's structure, such as the A-pillar shown in Figure \ref{fig:static_occlusion}, and dynamic occlusions caused by transient obstructions from other road users, like the truck shown in Figure \ref{fig:dynamic_occlusion}. Such perceptual gaps undermine an autonomous system's situational awareness, severely impairing its decision-making and path-planning capabilities and thereby elevating collision risk.

While onboard multi-sensor fusion systems can mitigate static occlusions, they remain vulnerable to dynamic occlusions in dense traffic \cite{kumar2023surround}. 
Vehicle-to-Everything (V2X) cooperative perception presents a promising solution by enabling information sharing that transcends the line-of-sight limitations of a single vehicle \cite{caillot2022survey, garcia2021tutorial}. 
However, its widespread adoption is impeded by a fundamental research gap: the absence of a standardized framework to quantify occlusion-induced risk and systematically evaluate the safety benefits of V2X under varying deployment parameters, such as the penetration rate of connected vehicles. 
This gap remains a primary obstacle to the technology's large-scale validation and deployment.

In the previous studies, occlusion risk assessment was confined to a single-vehicle perspective. This forced methods to model latent threats by, for example, generating 'virtual obstacles' and then quantifying risk using techniques like reachability analysis \cite{koschi2020set, park2023occlusion, chen2025occlusion} based on local inferences \cite{yu2019occlusion}. The lack of ground-truth data in occluded regions inherently constrained the efficacy of these methods. Consequently, they had to operate on assumptions about unseen agents rather than empirical observations, which limited their ability to model complex, real-world risks. However, the recent emergence of large-scale datasets, particularly from aerial views and cooperative perception \cite{xu2022drone,xu2023v2v4real}, is now providing this crucial ground truth. These resources provide the high-fidelity data required to precisely quantify occlusion risk, with aerial datasets, in particular, offering near-complete visibility into previously obscured traffic dynamics.

Building on this data-rich foundation, prior research has evaluated the benefits of cooperative perception primarily using perception-layer benchmarks, with Average Precision (AP) in detecting occluded targets as the predominant metric \cite{song2025traf, 10919014, xu2022v2x}. 
Although more targeted metrics, such as the occlusion recovery rate, have been proposed \cite{11096563}, they remain perception-layer metrics that are not necessarily indicative of a quantifiable improvement in vehicle safety. A critical limitation persists: they quantify the ability to detect an occluded object but fail to establish a direct link between this perceptual improvement and the resulting reduction in downstream collision risk.

To bridge this gap, some studies have shifted their focus to safety-layer metrics, such as collision probability in simulated hazardous scenarios \cite{10.1145/3498361.3538925,cui2022coopernaut,deng2019cooperative,wang2024deepaccident}. 
However, their reliance on oversimplified, low-density traffic conditions limits their generalizability, for example, metrics like Maximum Tracking Loss (MTL) are constrained by a narrow focus on Vulnerable Road Users (VRUs) and coarse risk quantification \cite{Wolff2024}.

In response to these challenges, we introduce a comprehensive framework to quantify, analyze, and mitigate occlusion-induced collision risk. We develop a novel risk metric based on temporal accumulation, employ it in a large-scale statistical analysis to characterize real-world hazards, and ultimately apply the framework to evaluate and optimize the deployment of V2X cooperative perception.

This principle of temporal accumulation is central to our framework's contribution. By integrating instantaneous risk over the entire duration an agent is occluded, our metric offers a holistic assessment that captures both acute, short-term dangers and the cumulative threat from low-intensity, long-duration events, such as a vehicle lingering in a blind spot. 
This instantaneous risk is modeled by weighting relative kinematics and spatial dynamics to measure latent collision potential across diverse traffic contexts accurately. Unlike safety-agnostic benchmarks such as AP that penalize all failures equally, RTL weights hazards based on their kinematic context. By shifting the focus from sensing performance to safety redundancy, RTL translates “what is seen” into “what is at stake”, establishing a physically grounded link between perception and system-level safety.

We rigorously validated the framework’s efficacy and generalizability through extensive experiments on two diverse datasets: SIND \cite{xu2022drone} and the Waymo Open Dataset \cite{sun2020scalability}, which focus on urban intersections and urban roads, respectively.
Our analysis systematically characterizes hazardous occlusion scenarios involving vehicles and vulnerable road users in complex urban settings.
This process serves a dual purpose: it confirms the robustness of our risk metric and establishes a curated collection of representative real-world hazardous cases to support the development of next-generation cooperative perception systems.

Leveraging this validated framework, we further investigate deployment strategies for cooperative perception, scrutinizing the nonlinear relationship between connected vehicle penetration rates and risk mitigation. 
Central to this analysis is a comparison between the conventional symmetric vehicle-to-vehicle (V2V) communication paradigm and a novel asymmetric one we propose, where even non-connected vehicles can receive safety broadcasts. 
Our quantitative results demonstrate that the asymmetric paradigm achieves equivalent or superior safety gains at significantly lower penetration rates, establishing it as a more cost-effective deployment strategy. These findings provide a strategic roadmap for the phased, economically viable rollout of V2X technology.

The main contributions of this work are summarized as follows:
\begin{itemize}
\item We propose the Risk of Tracking Loss (RTL), a universal occlusion risk metric based on a temporal accumulation principle. The RTL provides a holistic assessment by capturing both high-intensity, short-term threats and prolonged, low-intensity exposures to build a more comprehensive risk profile.
\item We conduct a large-scale statistical analysis of real-world occlusion risk using diverse, high-fidelity datasets. This characterization of hazardous scenarios offers crucial empirical insights into how risk manifests in complex traffic, informing system design and validation.
\item We apply our framework to evaluate V2X deployment strategies and identify a cost-effective implementation pathway. We demonstrate that a novel asymmetric communication paradigm can achieve superior safety outcomes at lower penetration rates, offering a strategic roadmap for real-world deployment.
\end{itemize}

\section{{Related work}}

\subsection{Occlusion-Induced Risk Quantification}
Quantifying occlusion-induced risk is a fundamental challenge in autonomous driving. Initial research addressed this from a single-vehicle perspective, modeling latent threats by generating virtual obstacles in occluded areas. The associated risk was quantified using techniques ranging from early spatio-temporal probability predictions \cite{yu2019occlusion, mcgill2019probabilistic} to more sophisticated methods such as formal set-based reachability analysis \cite{koschi2020set}, simplified reachability models \cite{park2023occlusion}, and spatio-temporal projection \cite{chen2025occlusion}. A common limitation of these approaches is their operation without ground-truth data for occluded regions. Consequently, they rely on assumptions regarding the presence of a virtual obstacle rather than on empirical observation, which hinders their ability to model the complex and stochastic nature of real-world risk.

The recent availability of large-scale, high-fidelity datasets from aerial and cooperative perception platforms has provided access to ground-truth information for these occluded regions \cite{xu2022drone,xu2023v2v4real}. However, existing works have yet to fully leverage this data. For instance, while Zhang et al. \cite{zhang2023occlusion} utilized V2X messages to refine the appearance probability of virtual obstacles, the framework's single-vehicle viewpoint precludes the full utilization of ground-truth data. Similarly, metrics like MTL \cite{Wolff2024}, despite using ground-truth data for occlusion analysis, are specialized for VRUs, limiting their generalizability. In contrast, our work introduces a universal, cumulative risk metric. This metric is validated through a large-scale statistical analysis of real-world hazardous occlusion events, establishing an empirical foundation for the design and validation of safer cooperative perception systems.

\subsection{Cooperative Perception Safety Benefits}

The safety benefits of cooperative perception have been predominantly evaluated indirectly, using perception-layer benchmarks. Metrics such as AP in detecting occluded targets are commonly employed to demonstrate the advantages of shared sensor data \cite{10919014, song2025traf, xu2022v2x}. While more specialized metrics like the occlusion recovery rate have also been proposed to quantify improvements more directly \cite{11096563}, a critical limitation persists across these approaches: they fail to establish a direct link between enhanced perception and a tangible reduction in downstream collision risk. To address this limitation, some research has shifted towards safety-layer evaluations. These studies typically rely on manually designed hazardous scenarios within simulators, assessing safety benefits through surrogate metrics, such as collision rate \cite{10.1145/3498361.3538925,cui2022coopernaut,deng2019cooperative,wang2024deepaccident}. However, the core limitation of this approach is its dependence on hand-crafted cases. Such simulations often feature oversimplified, low-density traffic conditions and a narrow set of predefined encounter types. Consequently, they fail to capture the statistical diversity and complexity of real-world hazards, which constrains the generalizability of their conclusions and underscores the gap between demonstrated perceptual gains and verifiable safety improvements.

\section{Method}
\label{method}
\subsection{Methodology Overview}
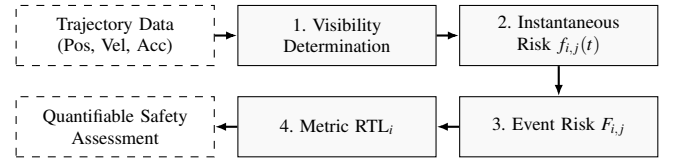
\begin{figure}[t]
\centering
\begin{tikzpicture}[
    node distance=0.4cm and 0.3cm, 
    auto,
    process/.style={rectangle, text width=2.4cm, minimum height=0.8cm, align=center, draw=black, fill=gray!5, font=\scriptsize},
    io/.style={rectangle, text width=2.4cm, minimum height=0.8cm, align=center, draw=black, fill=white, font=\scriptsize, dashed},
    arrow/.style={thick, -{Latex[length=1.5mm]}, font=\tiny}
]
    \node (input) [io] {Trajectory Data\\(Pos, Vel, Acc)};
    \node (step1) [process, right=of input] {1. Visibility Determination};
    \node (step2) [process, right=of step1] {2. Instantaneous Risk $f_{i,j}(t)$};

    \node (step3) [process, below=of step2] {3. Event Risk $F_{i,j}$};
    \node (step4) [process, left=of step3] {4. Metric $\text{RTL}_i$};
    \node (output) [io, left=of step4] {Quantifiable Safety\\Assessment};

    \draw [arrow] (input) -- (step1);
    \draw [arrow] (step1) -- (step2);
    \draw [arrow] (step2) -- (step3); 
    \draw [arrow] (step3) -- (step4); 
    \draw [arrow] (step4) -- (output);
\end{tikzpicture}
\caption{The RTL computational pipeline illustrating the hierarchical flow from trajectory inputs to system-level safety evaluation.
}
\label{fig:method_overview}
\end{figure}

The computational framework for the RTL metric transitions from raw trajectory data to quantifiable safety indicators through a structured four-stage sequence as shown in Figure \ref{fig:method_overview}. This pipeline establishes a unified notation system and physical assumptions to ensure consistency throughout the risk assessment process:
\begin{itemize}
    \item Visibility Determination: This stage evaluates the line-of-sight (LoS) and field-of-view (FoV) constraints for each road user. An agent $RU_j$ is classified as occluded if it remains within the perceptual range of the subject vehicle $RU_i$ but is visually obstructed by infrastructure or other vehicles.
    \item Instantaneous Risk Estimation: For each identified occlusion, the framework calculates an instantaneous risk weight $P$ based on relative kinematics. This weight defines the risk intensity $f_{i,j}(t)$ at each discrete time frame, representing the immediate threat level.
    \item Event-level Risk Evaluation: The framework integrates the risk intensity $f_{i,j}(t)$ over continuous occluded intervals. This yields the area $F_{i,j}$, which captures the total sustained risk from the onset of the perceptual gap to its resolution.
    \item System-level Risk Evaluation: The final stage aggregates interaction-level risks to determine the $\text{RTL}_i$ for each subject vehicle. By applying a worst-case formulation to select the maximum $F_{i,j}$ among all interacting participants, the metric identifies the most critical hazard in the environment.
\end{itemize}

This hierarchical approach ensures that the assessment is grounded in physical interactions while capturing the cumulative nature of safety redundancy depletion. Detailed mathematical definitions for each stage follow in the subsequent subsections. To facilitate the understanding of the mathematical formulations and the risk assessment framework presented herein, the key symbols and notations used throughout this paper are summarized in Table \ref{tab:symbols}.

\begin{table}[t]
\caption{Summary of Key Notations}
\label{tab:symbols}
\centering
\footnotesize %
\renewcommand\arraystretch{1.1} %

\begin{tabular*}{\linewidth}{@{\extracolsep{\fill}} l l}
\toprule
\textbf{Symbol} & \textbf{Description} \\
\midrule

\multicolumn{2}{l}{\textit{\textbf{Risk Assessment Framework}}} \\
$RU_i, RU_j$ & Subject road user and interacting road user \\
$t, T$ & Discrete time frame index and total time horizon \\
$f_{i,j}(t)$ & Instantaneous risk intensity imposed by the interacting road user \\
$A_k$ & Risk integral of a continuous occlusion event \\
$F_{i,j}$ & Maximal sustained risk among all occlusion events \\
$RTL_i$ & Risk of Tracking Loss metric for the subject road user \\
\midrule

\multicolumn{2}{l}{\textit{\textbf{Kinematics and Interaction Parameters}}} \\
$\mathbf{p}, \mathbf{v}$ & Position and velocity vectors of road users \\
$d$ & Euclidean distance between road users \\
$\Delta v$ & Magnitude of the relative velocity vector \\
$v_{rel}$ & Radial relative velocity projected onto the distance vector \\
$\theta$ & Angle between velocity vectors of interacting road users \\
$P$ & Normalized instantaneous risk weight based on kinematics \\
$k$ & General risk coefficient governing risk decay \\
$k_{\{\cdot\}}$ & Specific risk coefficients: $k_{static}, k_{dynamic}, k_{overlap}, k_{no\text{-}overlap}$ \\
$I_{over}$ & Indicator function for predicted reachable set overlap \\
$I_{side}$ & Indicator function for side-on interaction configuration \\
$M$ & Safety margin for reachable set expansion \\
$D$ & Prediction horizon for trajectory projection \\
\midrule

\multicolumn{2}{l}{\textit{\textbf{Statistical Evaluation Metrics}}} \\
$C_{QD}$ & Coefficient of Quartile Dispersion \\
$CV_{MAD}$ & Coefficient of Variation based on Median Absolute Deviation \\
$Q_1, Q_3$ & First and third quartiles of the distribution \\
$\text{MAD}$ & Median Absolute Deviation \\

\bottomrule
\end{tabular*}
\end{table}

\subsection{Metric Definition}

\begin{figure}[t]
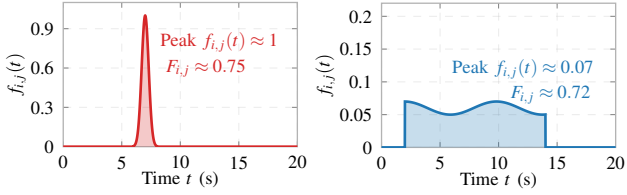

    \centering
    \noindent
    {%
     \captionsetup[subfigure]{justification=raggedright, singlelinecheck=false}
     \subfloat[High-intensity instantaneous risk.]{%
        \begin{minipage}{0.53\linewidth}%
            \raggedright
            \input{tikztex/case1_subplot}%
            \label{fig:high_intensity}
        \end{minipage}%
     }%
    }%
    \hspace{-30pt}%
    {%
     \captionsetup[subfigure]{justification=raggedleft, singlelinecheck=false}
     \subfloat[Low-intensity cumulative risk.]{%
        \begin{minipage}{0.53\linewidth}%
            \raggedleft
            \input{tikztex/case2_subplot}%
            \label{fig:low_intensity}
        \end{minipage}%
     }%
    }%
    \caption{Demonstration of two risk profiles captured by the proposed metric.}
    \label{fig:two_subfigures}
\end{figure}

We define a metric to quantify the risk associated with perception blind spots. Specifically, for two road users, $\text{RU}_i$ and $\text{RU}_j$, their instantaneous risk $f_{i,j}(t)$  at time $t$  is defined as:
\begin{equation}
f_{i,j}(t) = 
\begin{cases}
0, & \text{if } \text{RU}_j \text{ can be observed by } \text{RU}_i \\
P, & \text{otherwise}
\end{cases}
\end{equation}

When $\text{RU}_j$ can perceive $\text{RU}_i$, we assume $\text{RU}_j$ has sufficient decision-making capacity to avoid potential conflicts with $\text{RU}_i$. Under this condition of comprehensive sensory data and effective decision-making, the instantaneous risk is deemed negligible. Consequently, the function $f_{i,j}(t)$ is defined as zero.

Conversely, if $\text{RU}_j$ cannot perceive $\text{RU}_i$, this lack of perceptual information hinders its ability to make timely and effective decisions, creating a potential risk. This risk intensity is quantified by a non-zero value $P$. The value of $P$ is determined by a rule based on the motion states of both road users, which enhances the metric's practicality and discriminative power. Therefore, $f_{i, j}(t)$ quantifies the instantaneous risk $\text{RU}_j$ poses to $\text{RU}_i$ stemming from $\text{RU}_j$'s inability to acquire $\text{RU}_i$'s state information.

The risk function $f_{i,j}(t)$ may be positive over several disjoint time intervals, forming enclosed regions with the time axis. The area of each such region is:
\begin{equation}
A_k = \int_{t_a}^{t_b} f_{i,j}(t) , dt
\end{equation}
where the interval $[t_a, t_b]$ satisfies:
\begin{enumerate}
\item \label{cond:zero_boundary}
$f_{i,j}(t_a) = f_{i,j}(t_b) = 0$
\item \label{cond:positive_interval}
$\forall t \in (t_a, t_b)$, we have $f_{i,j}(t) > 0$
\end{enumerate}

We define $F_{i,j}$ as the maximum area among all such $n$ enclosed regions:
\begin{equation}
F_{i,j} = \max \{ A_1, A_2, \dots, A_n \}
= \max_{\substack{[t_a,t_b] \subseteq \mathbb{R} }} \int_{t_a}^{t_b} f_{i,j}(t)  dt
\end{equation}
where the maximization is over all intervals $\left[t_a, t_b\right]$ satisfying conditions \ref{cond:zero_boundary} and \ref{cond:positive_interval}.

In practice, time is discretized into a sequence of frames $t= 1, 2, \ldots, T$. The discrete formulation for $F_{i, j}$ is:
\begin{equation}
F_{i,j} = \max_k \sum_{t=t_k^{\text{start}}}^{t_k^{\text{end}}} f_{i,j}(t)
\end{equation}
where the maximization is over all contiguous time intervals $k$ (from $t_k^{\text{start}}$ to $t_k^{\text{end}}$) for which $f_{i,j}(t) > 0$, and $f_{i, j}(t)$ is the instantaneous risk at frame $t$.

The significance of $F_{i, j}$ is that it quantifies the total sustained risk imposed on $\text{RU}_i$ due to $\text{RU}_j$'s persistent inability to perceive it over a continuous interval. Each enclosed area represents a complete "risk event," capturing its full lifecycle from onset and accumulation to resolution. Taking the maximum value signifies a focus on the most severe risk event. Compared to traditional peak-based instantaneous risk methods, our proposed integral metric $F_{i, j}$ not only captures high-intensity, short-duration risk events (Figure \ref{fig:high_intensity}) but also accurately quantifies the cumulative effect of low-intensity, long-duration risks (Figure \ref{fig:low_intensity}). This enables a unified and comprehensive evaluation of different risk patterns, as shown in Figure \ref{fig:two_subfigures}.

We define the RTL for each road user $\text{RU}_i$ as:
\begin{equation}
\text{RTL}_i = \max_j F_{i,j}
\end{equation}

The significance of $\text{RTL}_i$ is that it characterizes the highest sustained risk level $\text{RU}_i$ faces from any single road user in the environment. By maximizing over all other traffic participants ${\text{RU}}_j$, this metric identifies the single most threatening interacting agent and its associated worst-case risk event for $\text{RU}_i$.

Physically, RTL represents the equivalent duration of a peak-intensity risk event. For example, an RTL of 100 ms signifies a cumulative impact equivalent to 0.1 s of exposure at the maximum weight ($P=1$). As an event-centric metric, RTL is calculated as the integral of risk intensity from its onset ($P>0$) to its resolution ($P=0$). This "risk impulse" formulation ensures that RTL captures the total depletion of safety redundancy, effectively equating acute, short-duration hazards with persistent, low-intensity threats. Based on statistical distributions, we categorize risk into three levels: low ($<$50 ms), medium (50–200 ms), and high ($>$200 ms) risk. Unlike raw occlusion time, RTL integrates motion states to characterize the dynamic severity of interactions.

The RTL metric supports both offline analysis and real-time implementation. It utilizes lightweight kinematic projections with a computational complexity of $O(N)$ for ego-vehicles. While this study employs a global $O(N^2)$ approach for statistics, the low overhead ensures feasibility for on-board assessment. As a cumulative metric, RTL is intended for long-term risk monitoring through infrastructure such as roadside units. It evaluates intersection safety profiles instead of triggering instantaneous emergency maneuvers.

\subsection{Risk Weight Parameter Design}
We model the risk weight $P$ as a function of the motion states and spatial configurations of the two traffic participants. Let $\mathbf{p}_v$ and $\mathbf{p}_u$ be their positions and $\mathbf{v}_v$ and $\mathbf{v}_u$ be their velocity vectors, respectively. The relative distance $d$ and relative speed $\Delta v$ are given by:
\begin{equation}
d = |\Delta \mathbf{p}| = |\mathbf{p}_v - \mathbf{p}_u|
\end{equation}
\begin{equation}
\Delta v = |\Delta \mathbf{v}| = |\mathbf{v}_v - \mathbf{v}_u|
\end{equation}
The risk weight $P$ is then formulated as:
\begin{equation}
P = \min\left(1, \max\left(0, \frac{k \cdot \Delta v}{d^{2}}\right)\right)
\end{equation}
This formulation clamps the risk value to the normalized range $[0, 1]$. The term $\frac{\Delta v}{d}$ is the inverse of Time to Collision (TTC) and represents a basic threat level. We modify this term to $\frac{\Delta v}{d^2}$ to better model the nonlinear attenuation of risk with increasing distance, analogous to an inverse-square law. The components of $P$, including distance and relative speed, serve as kinematic constraints that define the required braking buffer and reaction time. This design aligns with the safety-physics principles of Responsibility-Sensitive Safety\cite{shalev2017formal} (RSS), where risk intensity reflects the spatial-temporal proximity to the physical boundaries of safe interaction rather than being a purely empirical index.

The behavior of $P$ is governed by the coefficient $k$, which has dimensions of distance to ensure $P$ is dimensionless. The coefficient is assigned using a two-stage hierarchy, first classifying the interaction as static or dynamic:
\begin{equation}
k = 
\begin{cases}
k_{\rm static}, & \text{if } \|\mathbf{v}_v\| = 0 \text{ or } \|\mathbf{v}_u\| = 0 \\
k_{\rm dynamic}, & \text{otherwise}
\end{cases}
\label{eq:k_choice}
\end{equation}

To determine the specific values for $k_{\rm static}$ and $k_{\rm dynamic}$, we define several key interaction parameters:
\begin{itemize}
\item The radial relative velocity:
\begin{equation}
v_{\rm rel} = \frac{\Delta\mathbf{p} \cdot \Delta\mathbf{v}}{d}
\end{equation}
This scalar indicates if the agents are approaching ($v_{\rm rel} < 0$) or separating ($v_{\rm rel} > 0$).
\item A side-on interaction indicator, where $\theta = \angle(\mathbf{v}_v, \mathbf{v}_u)$ is the angle between velocity vectors:
\begin{equation}
I_{\rm side} = \mathbf{1}\{\theta \in (45^\circ, 135^\circ)\}
\end{equation}
\item A predicted overlap indicator:
\begin{equation}
I_{\rm over} = \mathbf{1}\{\text{Reachable sets overlap}\}
\end{equation}
\end{itemize}

The overlap indicator $I_{\rm over}$ is a binary flag that indicates whether the future reachable sets of both participants overlap. First, we predict the base reachable set using a constant acceleration linear motion model over a 0.6 s horizon. This duration maintains high fidelity between the model and actual vehicle motion while being long enough to capture imminent interaction risks. This duration aligns with the physiological lower bound of human reaction time \cite{green2000long,johansson1971drivers}, thereby characterizing hazards at the edge of human avoidability while maintaining the kinematic fidelity of the CA model.

This base prediction is then expanded to define a comprehensive safety volume. This expansion accounts for two factors: a lateral uncertainty proportional to the vehicle's width, using a coefficient of 0.05, to model sway; and a total safety margin. This margin is informed by traffic flow theory, such as the Boryankov model \cite{pan2025lateral}, which suggests a 0.7 m minimum lateral safety distance. We adopt this and add an additional 0.3 m conservative safety buffer, resulting in a 1.0 m total safety margin. $I_{\rm over}$ is set to $1$ if these final expanded sets of the two participants intersect. Due to the stochastic nature of pedestrian movement, we model their reachable set as a circle, with a radius determined by the predicted displacement and a safety margin.

To filter out non-hazardous interactions with stationary vehicles, such as parked cars, and focus on dynamic conflicts, the coefficient $k_{\rm static}$ is assigned a low value when either participant is stationary:
\begin{equation}
k_{\rm static} = 
\begin{cases}
0.05, & v_{\rm rel} < 0 \quad (\text{approaching})\\
0.01, & v_{\rm rel} \ge 0 \quad (\text{separating})
\end{cases}
\label{eq:k_static}
\end{equation}

\begin{figure}[t]
\centering
\begin{tikzpicture}[
  grow=down,
  level 1/.style={sibling distance=40mm, level distance=10mm},
  level 2/.style={sibling distance=22mm, level distance=10mm},
  level 3/.style={sibling distance=22mm, level distance=10mm},
  edge from parent/.style={draw,-latex},
  every node/.style={align=center, font=\small}
]

\node (root) {$I_{\rm over}=1?$}
  child { node {$I_{\rm side}=1?$}
    child { node {$k_{\rm dynamic}=3$}
      edge from parent node[pos=0.25, left] {yes}
    }
    child { node {$k_{\rm dynamic}=1$}
      edge from parent node[pos=0.25, right] {no}
    }
    edge from parent node[pos=0.25, left] {yes}
  }
  child { node {$v_{\rm rel} < 0$?}
    child { node {$I_{\rm side}=1?$}
      child { node {$k_{\rm dynamic}=0.4$}
        edge from parent node[pos=0.25, left] {yes}
      }
      child { node {$k_{\rm dynamic}=0.2$}
        edge from parent node[pos=0.25, right] {no}
      }
      edge from parent node[pos=0.25, left] {yes}
    }
    child { node {$k_{\rm dynamic}=0.01$}
      edge from parent node[pos=0.25, right] {no}
    }
    edge from parent node[pos=0.25, right] {no}
  };
\end{tikzpicture}
\caption{Decision logic for the coefficient $k_{\rm dynamic}$ in dynamic scenarios.}
\label{fig:decision_tree}
\end{figure}
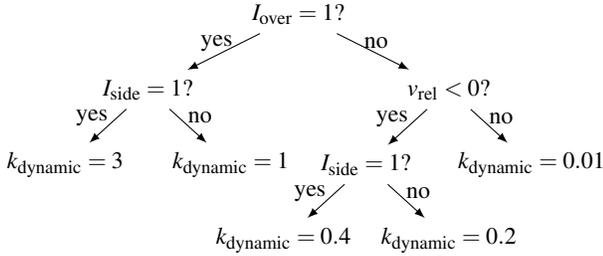
When both participants are in motion, the coefficient is $k_{\rm dynamic}$, assigned using the decision logic shown in Figure \ref{fig:decision_tree}. The rationale for this hierarchy is based on collision physics and kinematics. The $I_{\rm over}=1$ condition, indicating a predicted overlap of safety volumes, receives the highest weights as it represents the most imminent hazard \cite{manzinger2020using}. The higher weight for side impacts, $I_{\rm side}=1$, is justified by their increased severity; due to smaller crumple zones, side impacts transmit collision forces more directly, leading to a significantly higher risk of severe injury \cite{laberge2009occupant}. The $v_{\rm rel} < 0$ condition is also critical, as it ensures that non-negligible weights are assigned only when participants are approaching, while separating agents are correctly identified as low-risk.

This parameterization achieves a hierarchical risk weighting, with the specific coefficients calibrated on our validation dataset. The values were tuned to establish distinct orders of magnitude for different event types. This effectively separates the most hazardous scenarios, such as overlaps which result in a risk $P$ on the order of $10^{-1}$, from more benign interactions, such as simple approaches which yield a risk on the order of $10^{-2}$ or smaller.

 \section{Experiments}
\label{sec:sind}

\subsection{Experimental Setup}
Experiments are conducted on both the Sind and Waymo datasets, which encompass a variety of road scenarios. The simulation time step is synchronized with the trajectory data at a frequency of 10 Hz. Traffic participants are classified into vehicles (cars, trucks, buses) and VRUs (pedestrians, bicycles, motorcycles, tricycles). Our analysis focuses on two types of interaction pairs: vehicle-VRU (Veh-VRU) and vehicle-vehicle (Veh-Veh). Within a unified framework, our approach enables a consistent evaluation of perception capability, occlusion effects, and potential risks across both interaction types.

We model the perception and communication environment for each frame. All participants are assigned a perception range of 75 m. \cite{sun2020scalability}. Vehicle bodies are modeled as oriented ellipses for occlusion checks. Due to their smaller size, VRUs are considered non-occluding. To account for occlusion from static infrastructure, we identify road areas using dataset-provided map data, such as edge and center lines. All non-road regions are treated as occluding structures. A target is classified as visible only if it is within the observer's FoV and the LoS is unobstructed by either vehicles or static structures.

For cooperative perception, connected vehicles are sampled from the population based on a given penetration rate. We establish an adjacency network using a 200 m communication range that supports multi-hop message forwarding \cite{garcia2021tutorial}. All communication is assumed to be instantaneous, ignoring communication latency.

A worst-case formulation is adopted to identify critical safety bottlenecks and avoid the risk dilution inherent in mean-based metrics. This approach prioritizes the security of the most vulnerable participants, consistent with the methodology of MTL \cite{Wolff2024} and the safety principles of RSS \cite{shalev2017formal}.

\subsection{Risk Analysis}
\begin{figure*}[t]
    \centering
    \subfloat[Veh-Veh RTL and Traffic Volume in SIND]{
        \resizebox{0.31\linewidth}{!}{%
            \input{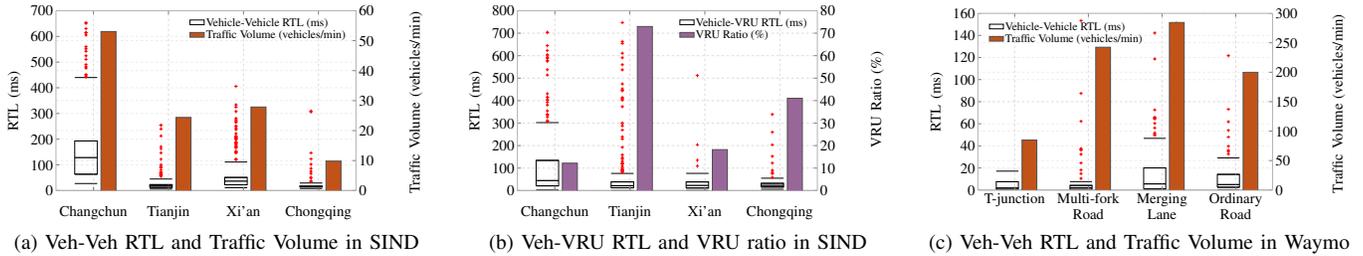}
        }
        \label{fig:combined_subfig1}
    }
    \hfill
    \subfloat[Veh-VRU RTL and VRU ratio in SIND]{
        \resizebox{0.31\linewidth}{!}{%
%
\definecolor{mycolor1}{rgb}{0.12941,0.12941,0.12941}%
\definecolor{boxcolor}{rgb}{0.40000,0.60000,0.750000}%
\definecolor{barcolor}{rgb}{0.60000,0.40000,0.60000}%
\definecolor{vrucolor}{rgb}{0.60000,0.40000,0.60000}%
\begin{tikzpicture}

\begin{axis}[%
width=5.0in,
height=3.0in,
at={(0.5in,0.5in)},
scale only axis,
xmin=0.5,
xmax=4.5,
xtick={1,2,3,4},
xticklabels={{Changchun},{Tianjin},{Xi'an},{Chongqing}},
xticklabel style={rotate=0, font=\LARGE, yshift=-15pt},
xlabel style={font=\LARGE\color{mycolor1}},
xlabel={},
ymin=0,
ymax=800,
ytick={0,100,200,300,400,500,600,700,800},
yticklabels={0,100,200,300,400,500,600,700,800},
yticklabel style={font=\LARGE},
ylabel style={font=\LARGE, yshift=5pt},
ylabel={RTL (ms)},
axis line style={solid, line width=0.8pt, color=gray!60}, 
xmajorgrids,
ymajorgrids,
yminorgrids,
axis y line*=left,
axis x line*=bottom,
grid style={dashed, opacity=0.3},
minor grid style={dashed, opacity=0.3},
legend style={legend cell align=left, align=left, fill=none, draw=none, font=\Large}
]

\addplot [color=black, forget plot]
  table[row sep=crcr]{%
0.75	302.135318137572\\
1.05	302.135318137572\\
};
\addplot [color=black, forget plot]
  table[row sep=crcr]{%
0.75	20.8885755474159\\
1.05	20.8885755474159\\
};
\addplot [color=black, forget plot]
  table[row sep=crcr]{%
0.75	134.122465573112\\
1.05	134.122465573112\\
};
\addplot [color=black, forget plot]
  table[row sep=crcr]{%
0.75	2.36021060032813\\
1.05	2.36021060032813\\
};
\addplot [color=black, forget plot]
  table[row sep=crcr]{%
0.75	20.8885755474159\\
0.75	134.122465573112\\
1.05	134.122465573112\\
1.05	20.8885755474159\\
0.75	20.8885755474159\\
};

\addplot [color=black, forget plot]
  table[row sep=crcr]{%
1.75	76.1884994148883\\
2.05	76.1884994148883\\
};
\addplot [color=black, forget plot]
  table[row sep=crcr]{%
1.75	12.2238874963709\\
2.05	12.2238874963709\\
};
\addplot [color=black, forget plot]
  table[row sep=crcr]{%
1.75	38.6004236381298\\
2.05	38.6004236381298\\
};
\addplot [color=black, forget plot]
  table[row sep=crcr]{%
1.75	0.00350244018073119\\
2.05	0.00350244018073119\\
};
\addplot [color=black, forget plot]
  table[row sep=crcr]{%
1.75	12.2238874963709\\
1.75	38.6004236381298\\
2.05	38.6004236381298\\
2.05	12.2238874963709\\
1.75	12.2238874963709\\
};

\addplot [color=black, forget plot]
  table[row sep=crcr]{%
2.75	76.4299256820476\\
3.05	76.4299256820476\\
};
\addplot [color=black, forget plot]
  table[row sep=crcr]{%
2.75	11.3423689666967\\
3.05	11.3423689666967\\
};
\addplot [color=black, forget plot]
  table[row sep=crcr]{%
2.75	38.4353666194423\\
3.05	38.4353666194423\\
};
\addplot [color=black, forget plot]
  table[row sep=crcr]{%
2.75	0.155631727195512\\
3.05	0.155631727195512\\
};
\addplot [color=black, forget plot]
  table[row sep=crcr]{%
2.75	11.3423689666967\\
2.75	38.4353666194423\\
3.05	38.4353666194423\\
3.05	11.3423689666967\\
2.75	11.3423689666967\\
};

\addplot [color=black, forget plot]
  table[row sep=crcr]{%
3.75	54.9619372927671\\
4.05	54.9619372927671\\
};
\addplot [color=black, forget plot]
  table[row sep=crcr]{%
3.75	15.6573984132893\\
4.05	15.6573984132893\\
};
\addplot [color=black, forget plot]
  table[row sep=crcr]{%
3.75	32.2621890182221\\
4.05	32.2621890182221\\
};
\addplot [color=black, forget plot]
  table[row sep=crcr]{%
3.75	5.72212101092705\\
4.05	5.72212101092705\\
};
\addplot [color=black, forget plot]
  table[row sep=crcr]{%
3.75	15.6573984132893\\
3.75	32.2621890182221\\
4.05	32.2621890182221\\
4.05	15.6573984132893\\
3.75	15.6573984132893\\
};

\addplot [color=black, line width=1.5pt, forget plot]
  table[row sep=crcr]{%
0.75	44.3591681810883\\
1.05	44.3591681810883\\
};

\addplot [color=black, line width=1.5pt, forget plot]
  table[row sep=crcr]{%
1.75	21.5033513690952\\
2.05	21.5033513690952\\
};

\addplot [color=black, line width=1.5pt, forget plot]
  table[row sep=crcr]{%
2.75	22.8067466803214\\
3.05	22.8067466803214\\
};

\addplot [color=black, line width=1.5pt, forget plot]
  table[row sep=crcr]{%
3.75	22.7341419109412\\
4.05	22.7341419109412\\
};

\addplot [color=mycolor1, only marks, mark=+, mark options={solid, fill=red, draw=red}, forget plot]
  table[row sep=crcr]{%
0.9	307.450677765374\\
0.9	308.942178727327\\
0.9	310.527757473806\\
0.9	327.63879100682\\
0.9	335.258181334463\\
0.9	339.681043488456\\
0.9	354.527139040824\\
0.9	380.636853552143\\
0.9	393.282143033657\\
0.9	402.032175874494\\
0.9	403.740902973327\\
0.9	414.294053833508\\
0.9	415.376489916991\\
0.9	430.646967478456\\
0.9	513.895463147574\\
0.9	537.553142104932\\
0.9	576.169295005749\\
0.9	577.193023720618\\
0.9	587.903936826573\\
0.9	593.096702999589\\
0.9	602.455894643906\\
0.9	603.02778961595\\
0.9	623.073603410663\\
0.9	644.503797336228\\
0.9	702.496069822836\\
0.9	703.692684591487\\
};

\addplot [color=mycolor1, only marks, mark=+, mark options={solid, fill=red, draw=red}, forget plot]
  table[row sep=crcr]{%
1.9	82.05005354266\\
1.9	82.6107436735863\\
1.9	89.4695800138511\\
1.9	90.1935094793186\\
1.9	90.6800207956729\\
1.9	92.1609533917352\\
1.9	98.116134560287\\
1.9	100.067673109991\\
1.9	108.014634087752\\
1.9	117.972804056112\\
1.9	119.595915460681\\
1.9	121.476924090864\\
1.9	129.843750339833\\
1.9	140.228805749018\\
1.9	147.139240833018\\
1.9	159.093315822304\\
1.9	192.711864947047\\
1.9	229.054778138966\\
1.9	239.047770303896\\
1.9	269.169813002428\\
1.9	287.960260984197\\
1.9	298.874734804404\\
1.9	300\\
1.9	373.305261310657\\
1.9	403.371346493942\\
1.9	403.455409061983\\
1.9	410.147401685515\\
1.9	413.589220892121\\
1.9	445.00957196467\\
1.9	445.364970231544\\
1.9	452.413101487609\\
1.9	474.152726222929\\
1.9	537.135417400521\\
1.9	559.886519758734\\
1.9	590.890688571782\\
1.9	610.728940043205\\
1.9	654.236158662741\\
1.9	662.980762519465\\
1.9	747.056093736062\\
};

\addplot [color=mycolor1, only marks, mark=+, mark options={solid, fill=red, draw=red}, forget plot]
  table[row sep=crcr]{%
2.9	108.868454665586\\
2.9	134.729243378613\\
2.9	203.574733903797\\
2.9	511.594659424708\\
};

\addplot [color=mycolor1, only marks, mark=+, mark options={solid, fill=red, draw=red}, forget plot]
  table[row sep=crcr]{%
3.9	58.47384088477\\
3.9	75.3854591084843\\
3.9	80.480311093754\\
3.9	89.9961458585224\\
3.9	122.26340408913\\
3.9	152.548122833083\\
3.9	203.320936911112\\
3.9	259.680496824993\\
3.9	338.807936828238\\
};

\end{axis}

\begin{axis}[%
width=5.0in,
height=3.0in,
at={(0.5in,0.5in)},
scale only axis,
xmin=0.5,
xmax=4.5,
xtick={1,2,3,4},
xticklabels={},
xticklabel style={rotate=0, font=\LARGE},
xlabel style={font=\LARGE},
xlabel={},
ymin=0,
ymax=80,
ytick={0,10,20,30,40,50,60,70,80},
yticklabels={0,10,20,30,40,50,60,70,80},
yticklabel style={font=\LARGE},
ylabel style={font=\LARGE, at={(axis cs:5.3,18)}, anchor=west},
ylabel={VRU Ratio (\%)},
axis y line*=right,
axis line style={solid, line width=0.8pt, color=gray!60}, 
xmajorgrids,
ymajorgrids,
yminorgrids,
grid style={dashed, opacity=0.3},
minor grid style={dashed, opacity=0.3},
legend style={legend cell align=left, align=left, fill=none, draw=none, font=\Large}
]

\addplot [fill=vrucolor, draw=white!30!black, forget plot] coordinates {
(1.1,0) (1.1,12.1879588839941) (1.3,12.1879588839941) (1.3,0)
};

\addplot [fill=vrucolor, draw=white!30!black, forget plot] coordinates {
(2.1,0) (2.1,72.9674796747967) (2.3,72.9674796747967) (2.3,0)
};

\addplot [fill=vrucolor, draw=white!30!black, forget plot] coordinates {
(3.1,0) (3.1,18.2033096926714) (3.3,18.2033096926714) (3.3,0)
};

\addplot [fill=vrucolor, draw=white!30!black, forget plot] coordinates {
(4.1,0) (4.1,41.0526315789474) (4.3,41.0526315789474) (4.3,0)
};

\node[centered, align=center, inner sep=0, font=\Large, color=white, font=\bfseries]
at (axis cs:1.2,15) {12.2\%};
\node[centered, align=center, inner sep=0, font=\Large, color=white, font=\bfseries]
at (axis cs:2.2,75) {73.0\%};
\node[centered, align=center, inner sep=0, font=\Large, color=white, font=\bfseries]
at (axis cs:3.2,20) {18.2\%};
\node[centered, align=center, inner sep=0, font=\Large, color=white, font=\bfseries]
at (axis cs:4.2,43) {41.1\%};

\addlegendimage{area legend, fill=white, draw=black}
\addlegendentry{Vehicle-VRU RTL (ms)};

\addlegendimage{area legend, fill=vrucolor, draw=white!30!black}
\addlegendentry{VRU Ratio (\%)};

\end{axis}

\end{tikzpicture}%
        }
        \label{fig:combined_subfig2}
    }
    \hfill
    \subfloat[Veh-Veh RTL and Traffic Volume in Waymo]{
        \resizebox{0.31\linewidth}{!}{%
%
\definecolor{mycolor1}{rgb}{0.12941,0.12941,0.12941}%
\definecolor{boxcolor}{rgb}{0.40000,0.60000,0.750000}%
\definecolor{barcolor}{rgb}{0.60000,0.40000,0.60000}%
\definecolor{trafficcolor}{rgb}{0.755000,0.32500,0.09800}%
\begin{tikzpicture}

\begin{axis}[%
width=5.0in,
height=3.0in,
at={(0.5in,0.5in)},
scale only axis,
xmin=0.5,
xmax=4.5,
xtick={1,2,3,4},
xticklabels={{T-junction},{Multi-fork\\Road},{Merging\\Lane},{Ordinary\\Road}},
xticklabel style={rotate=0, font=\LARGE, yshift=0pt, align=center},
xlabel style={font=\LARGE\color{mycolor1}},
xlabel={},
ymin=0,
ymax=160,
ytick={0,20,40,60,80,100,120,140,160},
yticklabels={0,20,40,60,80,100,120,140,160},
yticklabel style={font=\LARGE},
ylabel style={font=\LARGE, yshift=5pt},
ylabel={RTL (ms)},
axis line style={solid, line width=0.8pt, color=gray!60}, 
xmajorgrids,
ymajorgrids,
yminorgrids,
axis y line*=left,
axis x line*=bottom,
grid style={dashed, opacity=0.3},
minor grid style={dashed, opacity=0.3},
legend style={legend cell align=left, align=left, fill=none, draw=none, font=\Large}
]

\addplot [color=black, forget plot]
  table[row sep=crcr]{%
0.75	17.1079388392162\\
1.05	17.1079388392162\\
};
\addplot [color=black, forget plot]
  table[row sep=crcr]{%
0.75	0.595835410192547\\
1.05	0.595835410192547\\
};
\addplot [color=black, forget plot]
  table[row sep=crcr]{%
0.75	7.56101966861824\\
1.05	7.56101966861824\\
};
\addplot [color=black, forget plot]
  table[row sep=crcr]{%
0.75	0.0368930391032273\\
1.05	0.0368930391032273\\
};
\addplot [color=black, forget plot]
  table[row sep=crcr]{%
0.75	0.595835410192547\\
0.75	7.56101966861824\\
1.05	7.56101966861824\\
1.05	0.595835410192547\\
0.75	0.595835410192547\\
};

\addplot [color=black, forget plot]
  table[row sep=crcr]{%
1.75	7.63374762874957\\
2.05	7.63374762874957\\
};
\addplot [color=black, forget plot]
  table[row sep=crcr]{%
1.75	0.524192641554946\\
2.05	0.524192641554946\\
};
\addplot [color=black, forget plot]
  table[row sep=crcr]{%
1.75	4.19231026266546\\
2.05	4.19231026266546\\
};
\addplot [color=black, forget plot]
  table[row sep=crcr]{%
1.75	0.141198568675567\\
2.05	0.141198568675567\\
};
\addplot [color=black, forget plot]
  table[row sep=crcr]{%
1.75	0.524192641554946\\
1.75	4.19231026266546\\
2.05	4.19231026266546\\
2.05	0.524192641554946\\
1.75	0.524192641554946\\
};

\addplot [color=black, forget plot]
  table[row sep=crcr]{%
2.75	46.7976671451735\\
3.05	46.7976671451735\\
};
\addplot [color=black, forget plot]
  table[row sep=crcr]{%
2.75	0.949094175058619\\
3.05	0.949094175058619\\
};
\addplot [color=black, forget plot]
  table[row sep=crcr]{%
2.75	20.0438388136284\\
3.05	20.0438388136284\\
};
\addplot [color=black, forget plot]
  table[row sep=crcr]{%
2.75	0.101747239645081\\
3.05	0.101747239645081\\
};
\addplot [color=black, forget plot]
  table[row sep=crcr]{%
2.75	0.949094175058619\\
2.75	20.0438388136284\\
3.05	20.0438388136284\\
3.05	0.949094175058619\\
2.75	0.949094175058619\\
};

\addplot [color=black, forget plot]
  table[row sep=crcr]{%
3.75	29.0777699843544\\
4.05	29.0777699843544\\
};
\addplot [color=black, forget plot]
  table[row sep=crcr]{%
3.75	2.5440652195126\\
4.05	2.5440652195126\\
};
\addplot [color=black, forget plot]
  table[row sep=crcr]{%
3.75	14.1813792777438\\
4.05	14.1813792777438\\
};
\addplot [color=black, forget plot]
  table[row sep=crcr]{%
3.75	0.352382646031072\\
4.05	0.352382646031072\\
};
\addplot [color=black, forget plot]
  table[row sep=crcr]{%
3.75	2.5440652195126\\
3.75	14.1813792777438\\
4.05	14.1813792777438\\
4.05	2.5440652195126\\
3.75	2.5440652195126\\
};

\addplot [color=black, line width=1.5pt, forget plot]
  table[row sep=crcr]{%
0.75	1.88957639355229\\
1.05	1.88957639355229\\
};

\addplot [color=black, line width=1.5pt, forget plot]
  table[row sep=crcr]{%
1.75	2.0481533657897\\
2.05	2.0481533657897\\
};

\addplot [color=black, line width=1.5pt, forget plot]
  table[row sep=crcr]{%
2.75	5.55877490325251\\
3.05	5.55877490325251\\
};

\addplot [color=black, line width=1.5pt, forget plot]
  table[row sep=crcr]{%
3.75	4.94714978898046\\
4.05	4.94714978898046\\
};

\addplot [color=mycolor1, only marks, mark=+, mark options={solid, fill=red, draw=red}, forget plot]
  table[row sep=crcr]{%
1.9	10.467248453142\\
1.9	14.867444675588\\
1.9	18.1474661645446\\
1.9	24.4988282916809\\
1.9	32.3572770148891\\
1.9	36.1977637668003\\
1.9	36.7705035467736\\
1.9	37.4491834936193\\
1.9	62.2462139879274\\
1.9	87.5018601380758\\
1.9	153.389725825844\\
};

\addplot [color=mycolor1, only marks, mark=+, mark options={solid, fill=red, draw=red}, forget plot]
  table[row sep=crcr]{%
2.9	49.6654647935234\\
2.9	50.3303873019657\\
2.9	51.7576649584945\\
2.9	56.2379638901038\\
2.9	60.3619569063779\\
2.9	64.2389178486565\\
2.9	65.3562854701729\\
2.9	65.4435485105883\\
2.9	65.6291869219476\\
2.9	72.5444243576399\\
2.9	118.69951402487\\
2.9	142.218284142619\\
};

\addplot [color=mycolor1, only marks, mark=+, mark options={solid, fill=red, draw=red}, forget plot]
  table[row sep=crcr]{%
3.9	32.555378112773\\
3.9	34.9385321912496\\
3.9	35.9065805292807\\
3.9	39.6824890402794\\
3.9	47.9904509219675\\
3.9	53.9868505930738\\
3.9	61.8315414895656\\
3.9	73.1593418056797\\
3.9	121.674032338701\\
};

\end{axis}

\begin{axis}[%
width=5.0in,
height=3.0in,
at={(0.5in,0.5in)},
scale only axis,
xmin=0.5,
xmax=4.5,
xtick={1,2,3,4},
xticklabels={},
xticklabel style={rotate=0, font=\LARGE},
xlabel style={font=\LARGE},
xlabel={},
ymin=0,
ymax=300,
ytick={0,50,100,150,200,250,300},
yticklabels={0,50,100,150,200,250,300},
yticklabel style={font=\LARGE},
ylabel style={font=\LARGE, at={(axis cs:5.4,-10)}, anchor=west},
ylabel={Traffic Volume (vehicles/min)},
axis y line*=right,
axis line style={solid, line width=0.8pt, color=gray!60}, 
xmajorgrids,
ymajorgrids,
yminorgrids,
grid style={dashed, opacity=0.3},
minor grid style={dashed, opacity=0.3},
legend pos=north west,
legend style={legend cell align=left, align=left, fill=none, draw=none, font=\Large}
]

\addplot [fill=trafficcolor, draw=white!30!black, forget plot] coordinates {
(1.1,0) (1.1,84.9540085231416) (1.3,84.9540085231416) (1.3,0)
};

\addplot [fill=trafficcolor, draw=white!30!black, forget plot] coordinates {
(2.1,0) (2.1,242.42295860501) (2.3,242.42295860501) (2.3,0)
};

\addplot [fill=trafficcolor, draw=white!30!black, forget plot] coordinates {
(3.1,0) (3.1,284.841581196079) (3.3,284.841581196079) (3.3,0)
};

\addplot [fill=trafficcolor, draw=white!30!black, forget plot] coordinates {
(4.1,0) (4.1,200.000313557047) (4.3,200.000313557047) (4.3,0)
};

\node[centered, align=center, inner sep=0, font=\Large, color=white, font=\bfseries]
at (axis cs:1.2,90) {85.0};
\node[centered, align=center, inner sep=0, font=\Large, color=white, font=\bfseries]
at (axis cs:2.2,250) {242.4};
\node[centered, align=center, inner sep=0, font=\Large, color=white, font=\bfseries]
at (axis cs:3.2,290) {284.8};
\node[centered, align=center, inner sep=0, font=\Large, color=white, font=\bfseries]
at (axis cs:4.2,205) {200.0};

\addlegendimage{area legend, fill=white, draw=black}
\addlegendentry{Vehicle-Vehicle RTL (ms)};

\addlegendimage{area legend, fill=trafficcolor, draw=white!30!black}
\addlegendentry{Traffic Volume (vehicles/min)};

\end{axis}

\end{tikzpicture}%
        }
        \label{fig:combined_subfig3}
    }
    
    \caption{Correlation between RTL and traffic exposure metrics. Each subplot jointly visualizes distributional risk via box plots on the left Y-axis against a corresponding traffic exposure metric, such as Traffic Volume or VRU Ratio, via bar charts on the right Y-axis. The figure presents analyses for (a-b) Veh-Veh and Veh-VRU in the SIND dataset and (c) Veh-Veh in the Waymo dataset.}
    \label{fig:combined_figure}
\end{figure*}

To ensure comprehensive scenario diversity, we validated our metric using the SIND and Waymo Open datasets. The SIND dataset provides trajectories from intersections in four distinct Chinese cities: Tianjin, Changchun, Chongqing, and Xi'an. From the Waymo dataset, we selected four additional scenario types: T-shaped intersections, multi-fork roads, standard road segments, and merging lanes. Collectively, these scenarios feature diverse traffic flows and geometric layouts. Due to the limited and statistically insignificant number of VRUs in the Waymo dataset, our analysis of it was restricted to Veh-Veh interactions. As both datasets are derived from real-world data, we simulated realistic perception constraints by assuming all road users possess a 120° forward-facing FoV. The 120° FoV baseline reflects substantial inherent risks in non-cooperative environments, resulting in the wider RTL distribution (up to 700 ms) observed in Figure \ref{fig:combined_figure}.

\begin{figure*}[t]
    \centering
    \subfloat[Tianjin (Veh-Veh)]{\includegraphics[width=0.24\linewidth]{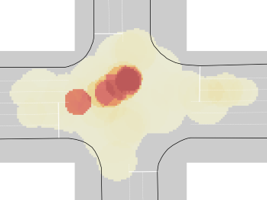}\label{subfig:2a}}\hfill
    \subfloat[Tianjin (Veh-VRU)]{\includegraphics[width=0.24\linewidth]{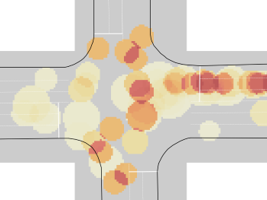}\label{subfig:2b}}\hfill
    \subfloat[Changchun (Veh-Veh)]{\includegraphics[width=0.24\linewidth]{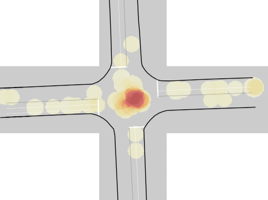}\label{subfig:2c}}\hfill
    \subfloat[Changchun (Veh-VRU)]{\includegraphics[width=0.24\linewidth]{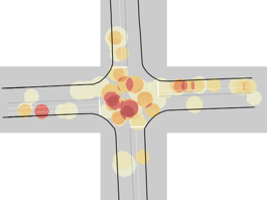}\label{subfig:2d}}
    
    \vspace{-0.5em}
    \subfloat[T-junction]{\includegraphics[width=0.24\linewidth]{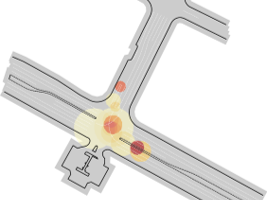}\label{subfig:2e}}\hfill
    \subfloat[Multi-fork Road]{\includegraphics[width=0.24\linewidth]{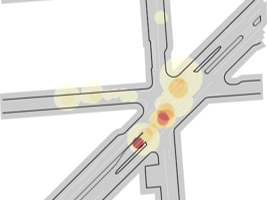}\label{subfig:2f}}\hfill
    \subfloat[Merging Lane]{\includegraphics[width=0.24\linewidth]{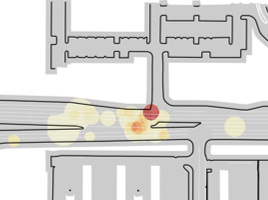}\label{subfig:2g}}\hfill
    \subfloat[Ordinary Road]{\includegraphics[width=0.24\linewidth]{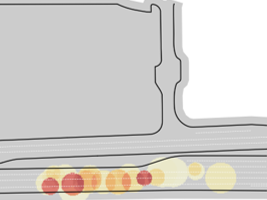}\label{subfig:2h}}  
    
    \vspace{0.5em}
    \includegraphics[scale=0.9]{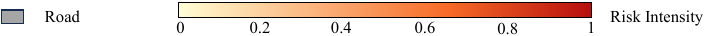}
    \vspace{-0.5em}
    \caption{Visualization of spatial risk aggregation heatmaps for scenarios from the SIND and Waymo datasets. The heatmaps are generated by accumulating RTL values of high-risk events at their peak-risk locations. Gray areas denote roads, and the color gradient indicates risk intensity, with warmer colors indicating higher accumulated RTL. The first row (a-d) displays intersection scenarios from the SIND dataset, comparing Veh-Veh and Veh-VRU at the Tianjin and Changchun intersections. The second row (e-h) presents various Veh-Veh interaction scenarios from the Waymo dataset, including a T-junction, a multi-fork road, a merging lane, and an ordinary road.}
    \label{fig:scenario_heatmaps}
\end{figure*}

\begin{figure*}[ht]
    \centering
    \subfloat[]{\includegraphics[width=0.325\linewidth]{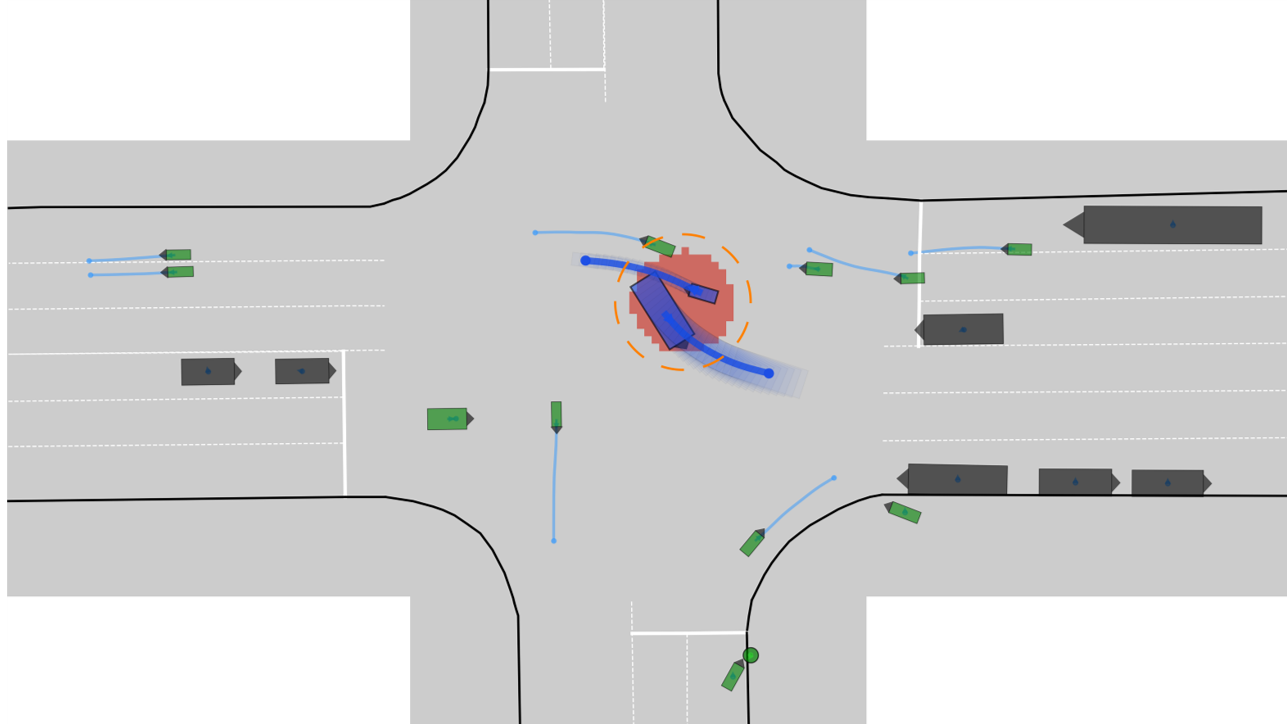}\label{subfig:3a}}\hfill
    \subfloat[]{\includegraphics[width=0.325\linewidth]{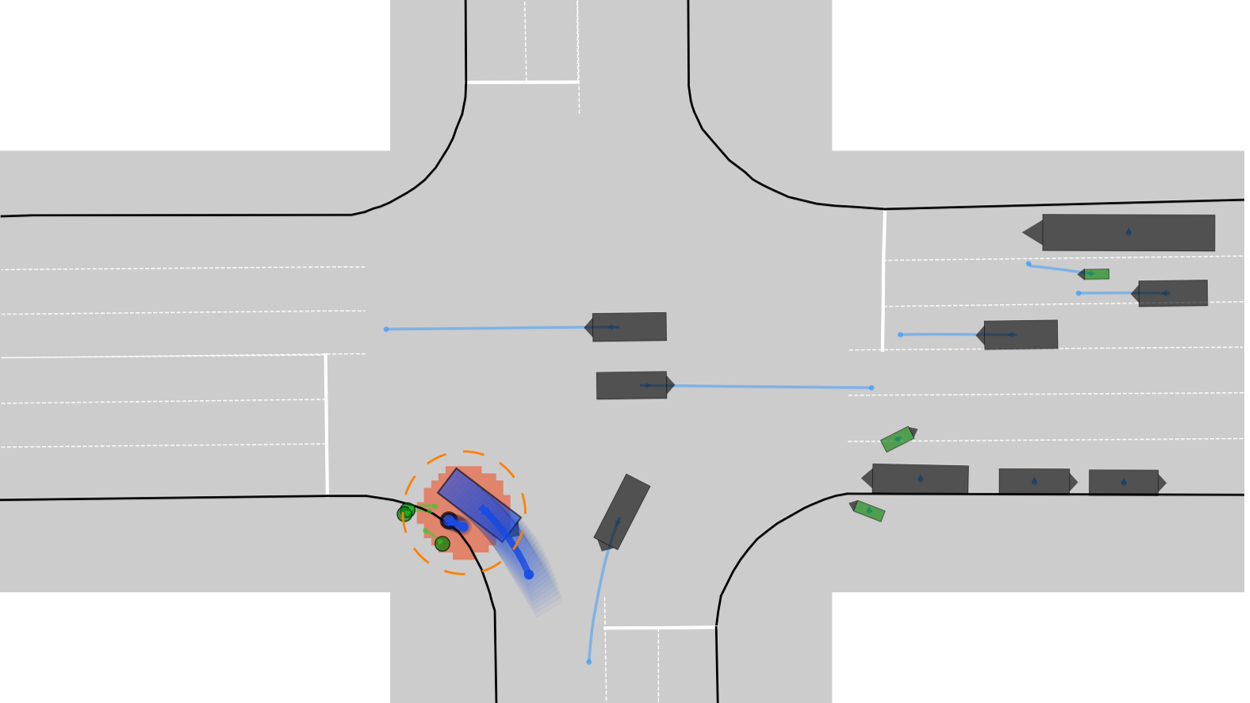}\label{subfig:3b}}\hfill
    \subfloat[]{\includegraphics[width=0.325\linewidth]{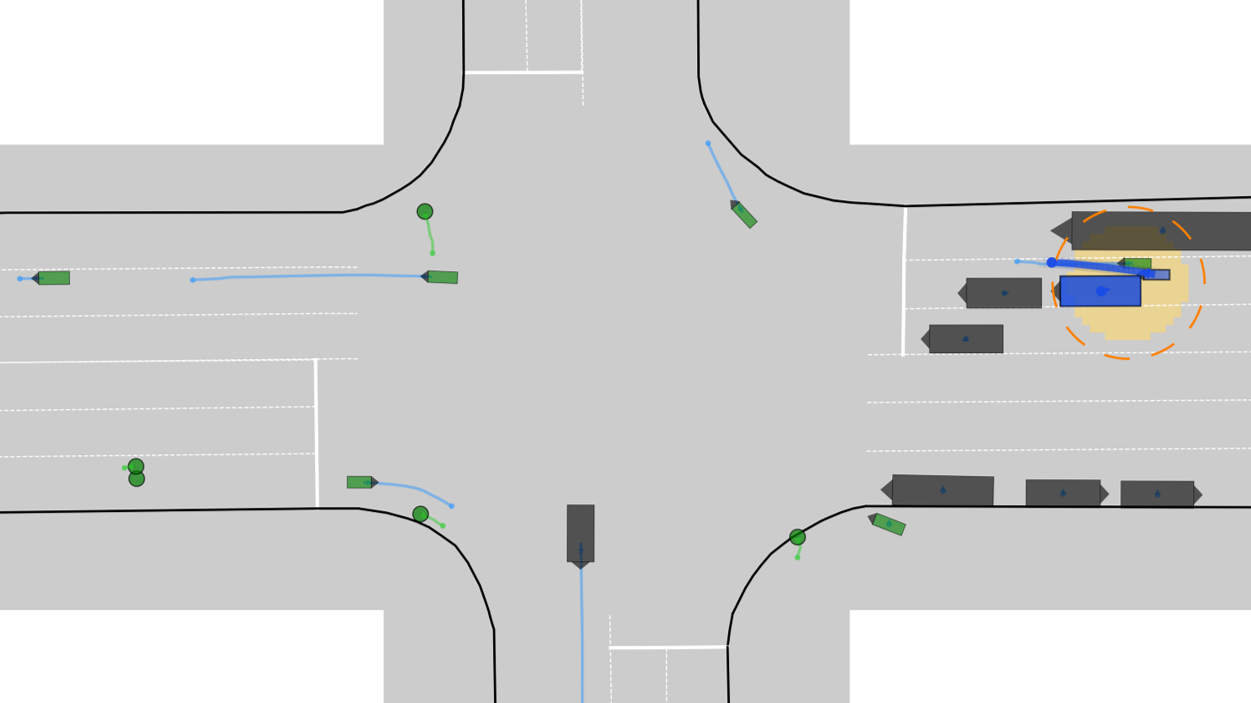}\label{subfig:3c}}
    
    \vspace{0em}
    \subfloat[]{\includegraphics[width=0.325\linewidth]{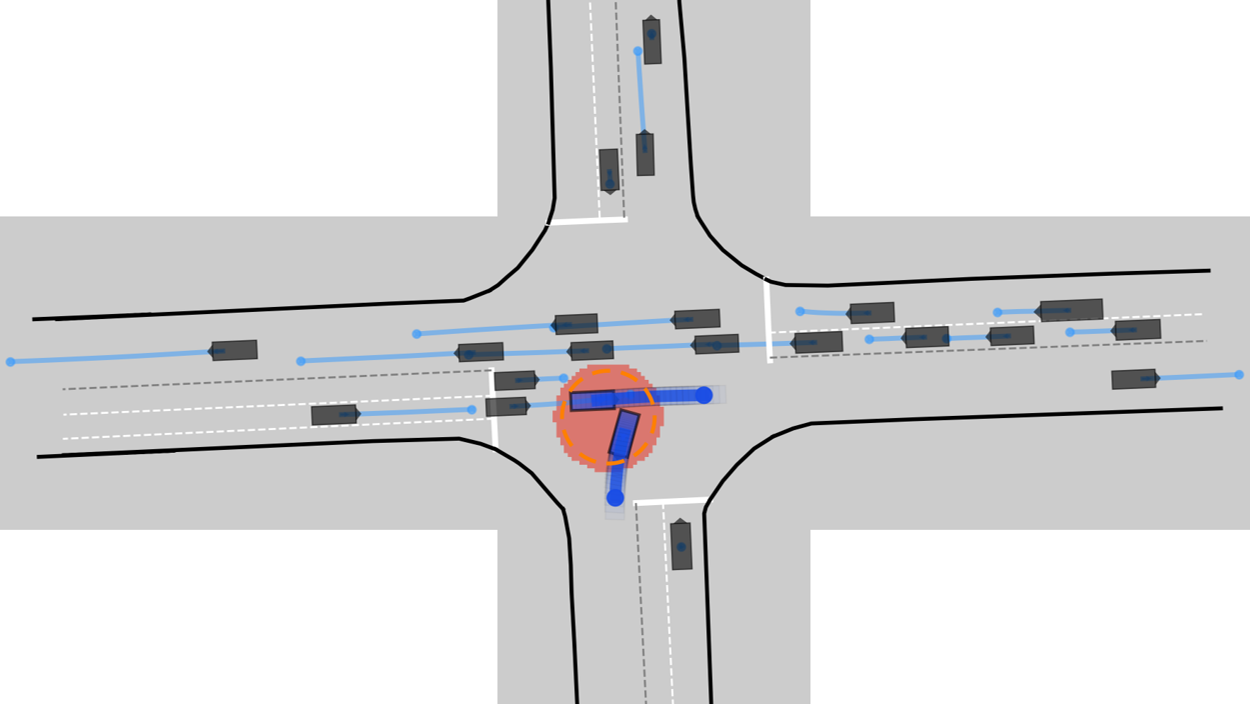}\label{subfig:3d}}\hfill
    \subfloat[]{\includegraphics[width=0.325\linewidth]{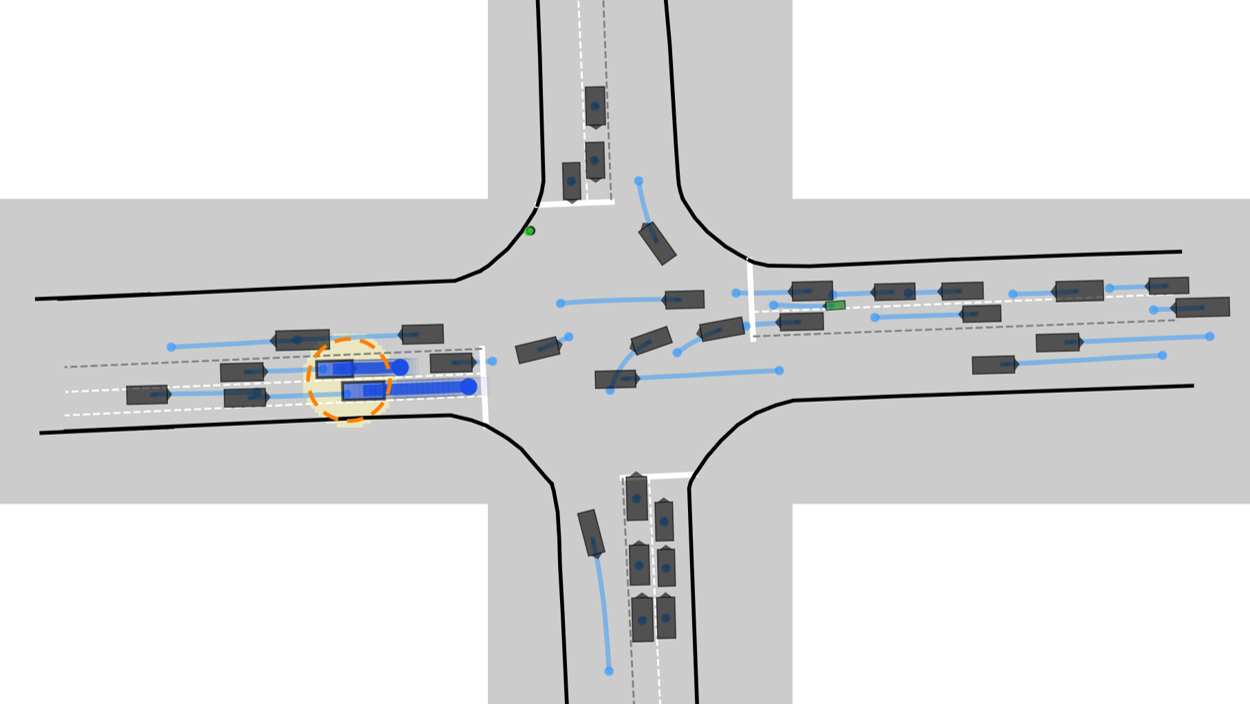}\label{subfig:3e}}\hfill
    \subfloat[]{\includegraphics[width=0.325\linewidth]{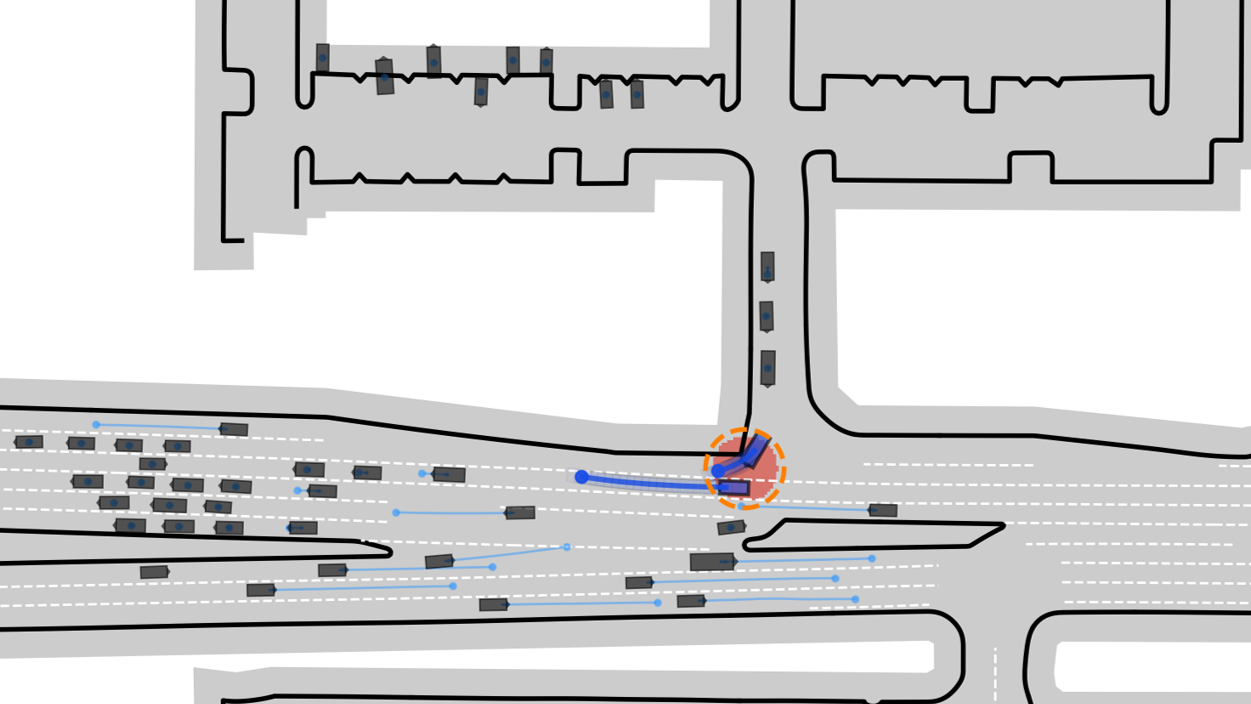}\label{subfig:3f}}
    \vspace{0em}
    \centering
    \includegraphics[scale=0.9]{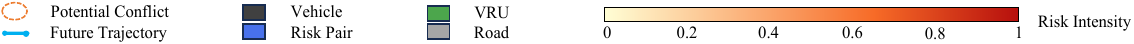}
    
    \caption{Representative risk events identified by the proposed metric across both datasets. In each subfigure, the circled pair highlights a risk interaction, with the circle's color indicating the single-frame risk value at the illustrated instant. (a) A non-motorized vehicle running a red light at an intersection. (b) A VRU within a vehicle's blind spot during a turning maneuver at a road edge convergence zone. (c) A bus occupies a non-motorized lane, forcing a VRU into the traffic flow. (d) A conflict during an unprotected left turn. (e) A prolonged car-following scenario exhibiting high cumulative risk. (f) A conflict arising from a lane merge.}
    \label{fig:risk_events}
\end{figure*}

\begin{table}[t]
\centering
\caption{Comparison of $C_{QD}$ and $CV_{MAD}$ Between MTL and RTL}
\label{tab:qcd_madcv_comparison}

\footnotesize %
\renewcommand\arraystretch{1.1} %

\begin{tabular*}{\linewidth}{@{\extracolsep{\fill}} l l c c c c}
\toprule
\multirow{2}{*}{Dataset} & \multirow{2}{*}{Scenario} & \multicolumn{2}{c}{$C_{QD}$} & \multicolumn{2}{c}{$CV_{MAD}$ (\%)} \\
\cmidrule(lr){3-4} \cmidrule(lr){5-6}
& & MTL & RTL & MTL & RTL \\ 
\midrule

\multirow{5}{*}{SIND} 
& Tianjin (Veh-Veh)   & 0.5722 & 0.5897 & 38.32 & 59.45 \\
& Tianjin (Veh-VRU)   & 0.3701 & 0.5160 & 37.76 & 58.31 \\
& Changchun (Veh-Veh) & 0.5232 & 0.7514 & 47.96 & 72.33 \\
& Changchun (Veh-VRU) & 0.2941 & 0.7259 & 27.81 & 65.99 \\
\cmidrule(lr){2-6}
& \textbf{SIND Mean}  & \textbf{0.4399} & \textbf{0.6458} & \textbf{37.96} & \textbf{64.02} \\
\midrule

\multirow{5}{*}{Waymo} 
& T-junction          & 0.7143 & 0.9395 & 80.60 & 100.00 \\
& Multi-fork Road     & 0.7105 & 0.8867 & 83.82 & 100.00 \\
& Merging Lane        & 0.5833 & 0.8377 & 72.31 & 97.09 \\
& Ordinary Road       & 0.5012 & 0.8816 & 67.76 & 98.12 \\
\cmidrule(lr){2-6}
& \textbf{Waymo Mean} & \textbf{0.6273} & \textbf{0.8861} & \textbf{76.12} & \textbf{98.80} \\

\bottomrule
\end{tabular*}
\end{table}

Figure \ref{fig:scenario_heatmaps} presents schematic diagrams of the scenarios, overlaid with risk heatmaps generated by our metric. To create these heatmaps, we first identified risk events with high RTL values. For each event, we marked the positions of the risk pair at the frame where the instantaneous risk value peaked. Second, circular masks were centered on these peak-risk positions. A regular grid was then superimposed on the map, and grid cells falling within any circle accumulated that event's corresponding RTL value. The final value of each grid cell thus represents the total accumulated risk at that spatial location. Finally, all grid values were normalized by the maximum value in that scenario and mapped to a color intensity scale to generate the heatmap.

To quantitatively evaluate the discriminative power of the proposed RTL metric, we conduct a comparative analysis against the MTL \cite{Wolff2024} using robust statistical dispersion measures. Specifically, we employ the Coefficient of Quartile Dispersion ($C_{QD}$) and the Coefficient of Variation based on Median Absolute Deviation ($CV_{MAD}$), defined as follows:
\begin{equation}
C_{QD} = \frac{Q_3 - Q_1}{Q_3 + Q_1}, \quad CV_{MAD} = \frac{1.4826 \cdot \text{MAD}}{\text{Median}}
\end{equation}
where $Q_1$ and $Q_3$ represent the first and third quartiles, respectively, and MAD denotes the median absolute deviation. These metrics are chosen for their resilience to extreme outliers, providing a more reliable assessment of a metric's ability to distinguish safety-critical events from baseline traffic distributions.

As summarized in Table \ref{tab:qcd_madcv_comparison}, RTL consistently exhibits higher $C_{QD}$ and $CV_{MAD}$ values across all evaluated scenarios in both SIND and Waymo datasets. For instance, within the SIND dataset, RTL achieves a mean $C_{QD}$ of 0.6458, significantly surpassing the 0.4399 observed for MTL. Similarly, the superior $CV_{MAD}$ values for RTL, such as 64.02\% compared to MTL's 37.96\% in the SIND mean, further highlight its heightened sensitivity to localized risk fluctuations. This increased statistical dispersion demonstrates that RTL effectively maps perceptual performance to system-level safety quantification. By establishing a physically grounded margin for identifying high-risk encounters, the metric avoids the risk dilution inherent in traditional benchmarks.

The heatmaps reveal that, in intersection scenarios, our metric identifies significantly higher risks within the intersection area compared to adjoining road segments. This concentration is primarily attributed to conflicts from unprotected left turns, which our metric effectively highlights. Furthermore, the metric also captures the more dispersed Veh-VRU, which stem from the stochastic motion patterns and frequent non-compliant behaviors of VRUs.

Notably, our metric also captures risks arising from environmental obstacles. In the Veh-VRU heatmap for the Tianjin scenario (Figure \ref{subfig:2b}), beyond the primary intersection risk zone, the results clearly delineate a narrow, elongated high-risk corridor along the motor vehicle lanes. This pattern accurately reflects a real-world situation where illegally parked buses occupy non-motorized lanes, forcing VRUs to navigate through gaps in motor vehicle traffic and thereby generating secondary risk patterns.

The box plots in Figure \ref{fig:combined_figure} summarize the statistical risk levels across the different scenarios, demonstrating a strong consistency with observed traffic conditions. The statistical results reveal two key trends. First, the relative risk ranking of both Veh-Veh and Veh-VRU conflicts across different intersections generally correlates with their respective total traffic volumes; higher volume intersections consistently exhibit higher overall risk levels. Second, the disparity between Veh-Veh and Veh-VRU levels within a single intersection is primarily governed by its VRU ratio. Intersections with a smaller VRU ratio exhibit a more significant reduction in Veh-VRU relative to their baseline Veh-Veh.

The Tianjin scenario provides excellent validation of this two-part logic in Figures \ref{fig:combined_subfig1} and \ref{fig:combined_subfig2}. Its low traffic density results in a correspondingly low baseline Veh-Veh, which aligns with the first trend. However, its exceptionally high VRU proportion, combined with a mixed traffic signal scheme that encourages non-compliance, prevents a drop in VRU-related risk. This results in prominent Veh-VRU interaction risks that are disproportionately high compared to its low vehicle traffic. This finding validates our metric's sensitivity to both aggregate traffic volume and specific contextual factors, notably the VRU ratio.

In the Waymo dataset, the risk levels identified by our metric also show a strong correlation with traffic density, as shown in Figure \ref{fig:combined_subfig3}. On multi-fork roads, however, the average risk level does not directly correspond to high traffic volume, a finding we attribute to the significant presence of parked vehicles along the roadside. Nonetheless, the peak risk values in these scenarios still consistently align with dense traffic conditions.

A key distinction between the datasets must be noted. Each SIND scenario contains 12,000 frames of data, whereas each Waymo scenario provides only 200. This disparity leads to limited interaction durations and a smaller observed vehicle population in the Waymo data. Furthermore, the geographical scale of the Waymo maps substantially exceeds that of SIND. Consequently, even with a higher vehicle count per minute, the observed aggregate risk levels in Waymo scenarios remain markedly lower than those in the SIND dataset.

Finally, the proposed metric effectively identifies diverse, specific risk events, as illustrated by the typical scenarios in Figure \ref{fig:risk_events}. In the SIND dataset, frequent VRU violations result in scattered Veh-VRU conflicts. In contrast, vehicle operations are generally rule-compliant and constrained, exhibiting no lane-changing or U-turn behaviors. The primary conflicts arise from unprotected left turns. The Waymo dataset, however, presents a broader range of inter-vehicle conflicts, including complex lane-merging maneuvers. Notably, in Figure \ref{subfig:3e}, although the instantaneous risk value remains low, a sustained car-following maneuver accumulates significant risk. This demonstrates our metric's ability to assess persistent, long-duration potential conflicts, not just instantaneous high-risk actions. To ensure the validity of the proposed metric, RTL is kinematically linked to TTC through its risk intensity term $P$. This correlation is qualitatively supported by the near-miss scenarios presented in Figure \ref{fig:risk_events}, confirming that RTL effectively quantifies the erosion of safety margins and that its reduction reliably implies improved system-level safety.

\subsection{Impact of V2X Penetration on Risk Reduction}
Our proposed metric accurately reflects macro-level risks consistent with real-world traffic conditions and, at the micro-level, effectively identifies localized hazards, particularly those in perception blind spots. This dual capability demonstrates the metric's suitability for characterizing risks arising from perceptual limitations.

\begin{figure*}[t]
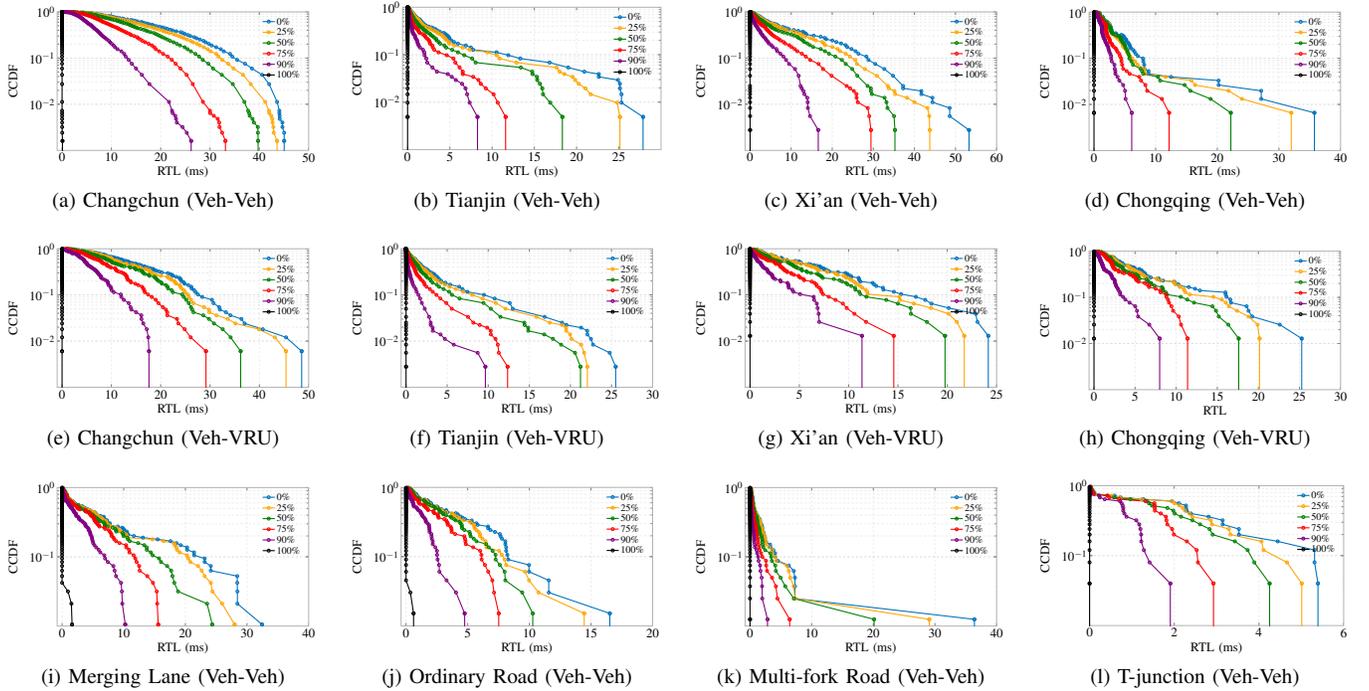

    \centering
    \subfloat[Changchun (Veh-Veh)]{
        \resizebox{0.23\linewidth}{!}{%
        \input{tikztex/changchun_veh_circle}
        }
    }
    \hfill
    \subfloat[Tianjin (Veh-Veh)]{
        \resizebox{0.23\linewidth}{!}{%
        \input{tikztex/tianjin_veh_circle}
        }
    }
    \hfill
    \subfloat[Xi'an (Veh-Veh)]{
        \resizebox{0.23\linewidth}{!}{%
        \input{tikztex/xian_veh_circle}
        }
    }
    \hfill
    \subfloat[Chongqing (Veh-Veh)]{
        \resizebox{0.23\linewidth}{!}{%
        \input{tikztex/chongqing_veh_circle}
        }
    }
    
    \vspace{0em}
    \subfloat[Changchun (Veh-VRU)]{
        \resizebox{0.23\linewidth}{!}{%
        \input{tikztex/changchun_vru_circle}
        }
    \label{subfig:e}}
    \hfill
    \subfloat[Tianjin (Veh-VRU)]{
        \resizebox{0.23\linewidth}{!}{%
        \input{tikztex/tianjin_vru_circle}
        }
    \label{subfig:f}}
    \hfill
    \subfloat[Xi'an (Veh-VRU)]{
        \resizebox{0.23\linewidth}{!}{%
        \input{tikztex/xian_vru_circle}
        }
    \label{subfig:g}}
    \hfill
    \subfloat[Chongqing (Veh-VRU)]{
        \resizebox{0.23\linewidth}{!}{%
        \input{tikztex/chongqing_vru_circle}
        }
    \label{subfig:h}}
    
    \vspace{0em}
    \subfloat[Merging Lane (Veh-Veh)]{
        \resizebox{0.23\linewidth}{!}{%
        \input{tikztex/huiru_veh_circle}
        }
    \label{subfig:i}}
    \hfill
    \subfloat[Ordinary Road (Veh-Veh)]{
        \resizebox{0.23\linewidth}{!}{%
        \input{tikztex/road_veh_circle}
        }
    \label{subfig:j}}
    \hfill
    \subfloat[Multi-fork Road (Veh-Veh)]{
        \resizebox{0.23\linewidth}{!}{%
        \input{tikztex/duochadao_veh_circle}
        }
    \label{subfig:k}}
    \hfill
    \subfloat[T-junction (Veh-Veh)]{
        \resizebox{0.23\linewidth}{!}{%
%
%
\definecolor{mycolor1}{rgb}{0.00000,0.44700,0.74100}%
\definecolor{mycolor2}{rgb}{1.00000,0.64710,0.00000}%
\definecolor{mycolor3}{rgb}{0.12941,0.12941,0.12941}%
\begin{tikzpicture}

\begin{axis}[%
width=4.9in,
height=2.7in,
at={(0.583in,0.675in)},
scale only axis,
xmin=0,
xmax=6,
xtick={ 0,  2, 4, 6},
xticklabel style={font=\LARGE},
xlabel style={font=\LARGE},
xlabel={RTL (ms)},
ymode=log,
ymin=0.01,
ymax=1,
ytick={0.1,1},
yticklabels={{$10^{-1}$},{$10^{0}$},{}},
yminorticks=true,
yticklabel style={font=\LARGE},
ylabel style={font=\LARGE, yshift=10pt},
ylabel={CCDF},
axis lines=box,
axis line style={solid, line width=0.8pt, color=gray!60}, 
xmajorgrids,
ymajorgrids,
yminorgrids,
grid style={dashed, opacity=0.3},
minor grid style={dashed, opacity=0.3},
legend style={legend cell align=left, align=left, fill=none, draw=none, font=\Large}
]
\addplot [color=mycolor1, line width=1.5pt, mark size=2.0pt, mark=o, mark options={solid, mycolor1}]
  table[row sep=crcr]{%
0.0349357240112996	0.96\\
0.0368930391032273	0.92\\
0.0619724154988197	0.88\\
0.0820034447282052	0.84\\
0.110407220517664	0.8\\
0.121229939587173	0.76\\
0.589992875556695	0.72\\
0.729147925082493	0.68\\
1.27307051314941	0.64\\
1.94585462333166	0.6\\
2.11028801291695	0.56\\
2.27732037040186	0.52\\
2.27732037040186	0.48\\
2.29382190546654	0.44\\
2.39236112985626	0.4\\
2.809874843361	0.36\\
3.14114833211635	0.32\\
3.14114833211635	0.28\\
3.52661908077877	0.24\\
3.52661908077877	0.2\\
4.44206170739774	0.16\\
5.31249390321265	0.12\\
5.31249390321265	0.08\\
5.39267329460891	0.04\\
5.39267329460891	0.04\\
5.39267329460891	0.0001\\
10.5	0.0001\\
};
\addlegendentry{0\%}

\addplot [color=mycolor2, line width=1.5pt, mark size=2.0pt, mark=o, mark options={solid, mycolor2}]
  table[row sep=crcr]{%
0.0349215852741365	0.96\\
0.0368930391032273	0.92\\
0.0614463993831011	0.88\\
0.0778039190639124	0.84\\
0.108353190690633	0.8\\
0.121229939587173	0.76\\
0.528201701533318	0.72\\
0.707787608813091	0.68\\
1.22583981691604	0.64\\
1.94305189490707	0.6\\
2.00212583773762	0.56\\
2.12008599941551	0.52\\
2.20281067083062	0.48\\
2.25077682163288	0.44\\
2.35710199023884	0.4\\
2.42665474389481	0.36\\
2.86417224122991	0.32\\
2.8894945173828	0.28\\
3.288550935648	0.24\\
3.3262207951733	0.2\\
4.05948241190437	0.16\\
4.10730618886819	0.12\\
4.6603676641698	0.08\\
5.00990350203149	0.04\\
5.00990350203149	0.04\\
5.00990350203149	0.0001\\
10.5	0.0001\\
};
\addlegendentry{25\%}

\addplot [color=black!50!green, line width=1.5pt, mark size=2.0pt, mark=o, mark options={solid, black!50!green}]
  table[row sep=crcr]{%
0.0263980455622287	0.96\\
0.0321149629777519	0.92\\
0.0557949085343516	0.88\\
0.0613333159184216	0.84\\
0.0971753867199587	0.8\\
0.098919806867632	0.76\\
0.421955727082382	0.72\\
0.577004321179594	0.68\\
0.96545742386136	0.64\\
1.36315751936166	0.6\\
1.40150289080305	0.56\\
1.4722126691706	0.52\\
1.83658027066752	0.48\\
1.9772782867093	0.44\\
2.00877271239455	0.4\\
2.16137745244684	0.36\\
2.42665474389481	0.32\\
2.6081418337868	0.28\\
2.83274886293435	0.24\\
2.90549731936564	0.2\\
3.42974976860858	0.16\\
3.73209250410248	0.12\\
3.88948587294449	0.08\\
4.25091184013668	0.04\\
4.25091184013668	0.04\\
4.25091184013668	0.0001\\
10.5	0.0001\\
};
\addlegendentry{50\%}

\addplot [color=red, line width=1.5pt, mark size=2.0pt, mark=o, mark options={solid, red}]
  table[row sep=crcr]{%
0.0225428266847122	0.96\\
0.0230179727600565	0.92\\
0.0443787761731655	0.88\\
0.0496625722965738	0.84\\
0.0815686062698431	0.8\\
0.0996698612899007	0.76\\
0.418458816737866	0.72\\
0.464644413854369	0.68\\
0.696497439197042	0.64\\
1.31970764996398	0.6\\
1.53009917425911	0.56\\
1.53627042596658	0.52\\
1.53776690312342	0.48\\
1.54111350135362	0.44\\
1.68019876256781	0.4\\
1.80323927103066	0.36\\
1.81536516650422	0.32\\
1.83534455095627	0.28\\
1.9485905951412	0.24\\
1.98984274902946	0.2\\
2.29630175793435	0.16\\
2.52871259424516	0.12\\
2.55920257215593	0.08\\
2.92596149602811	0.04\\
2.92596149602811	0.04\\
2.92596149602811	0.0001\\
10.5	0.0001\\
};
\addlegendentry{75\%}

\addplot [color=violet, line width=1.5pt, mark size=2.0pt, mark=o, mark options={solid, violet}]
  table[row sep=crcr]{%
0.00465445831756003	0.96\\
0.00621938994374317	0.92\\
0.0209561734579536	0.88\\
0.0295999522125885	0.84\\
0.0310590872796121	0.8\\
0.036339157151925	0.76\\
0.223984384911638	0.72\\
0.241529835887052	0.68\\
0.358057408147013	0.64\\
0.718704165609768	0.6\\
0.757480930361955	0.56\\
0.780482234357447	0.52\\
0.793435059745927	0.48\\
0.793717358386655	0.44\\
0.821516625317072	0.4\\
0.87201771797224	0.36\\
1.10241689731246	0.32\\
1.17278154298089	0.28\\
1.20104692750569	0.24\\
1.21792792127241	0.2\\
1.22021858935726	0.16\\
1.28109557674615	0.12\\
1.41716984173768	0.08\\
1.9067275867267	0.04\\
1.9067275867267	0.04\\
1.9067275867267	0.0001\\
10.5	0.0001\\
};
\addlegendentry{90\%}

\addplot [color=black, line width=1.5pt, mark size=2.0pt, mark=o, mark options={solid, black}]
  table[row sep=crcr]{%
0	0.96\\
0	0.92\\
0	0.88\\
0	0.84\\
0	0.8\\
0	0.76\\
0	0.72\\
0	0.68\\
0	0.64\\
0	0.6\\
0	0.56\\
0	0.52\\
0	0.48\\
0	0.44\\
0	0.4\\
0	0.36\\
0	0.32\\
0	0.28\\
0	0.24\\
0	0.2\\
0	0.16\\
0	0.12\\
0	0.08\\
0	0.04\\
0	0.04\\
0	0.0001\\
10.5	0.0001\\
};
\addlegendentry{100\%}

\end{axis}
\end{tikzpicture}%
        }
    \label{subfig:l}}
    
  \caption{Impact of Connectivity Penetration Rate on RTL. Subfigures~(a)--(d) present results from four cities in the SIND dataset, covering Veh-Veh interactions. Subfigures~(e)--(h) show Veh-VRU interactions from the same cities. Subfigures~(i)--(l) display four Veh-Veh scenarios from the Waymo dataset. Across all scenarios, safety benefits from increasing penetration rates far exceed linear growth, with improvements in the high-penetration regime vastly outperforming those at low penetration.}
    \label{fig:3x4_grid}
\end{figure*}

Building on this validation, we leverage the metric to investigate the risk-mitigation mechanisms of cooperative perception, which extend the effective perceptual range. We evaluate an autonomous vehicle system to quantify how the V2X penetration rate affects system-wide safety. To isolate this specific effect, we assume all vehicles are equipped with identical LiDAR sensors and model their perception as a uniform circular area. This configuration eliminates confounding variables from heterogeneous sensor fields of view, ensuring the results primarily reflect the influence of the penetration rate itself. The idealized 360° circular FoV setting explains the compressed RTL scale (mostly $<$50 ms) in Figure \ref{fig:3x4_grid} compared to Figure \ref{fig:combined_figure}.

\newcolumntype{Y}{>{\centering\arraybackslash}X}

\begin{table}[t]
\centering
\caption{Quantitative Comparison of Normalized RTL Reduction Across V2X Penetration Rates}
\label{tab:penetration}

\footnotesize
\renewcommand{\arraystretch}{1.1}
\setlength{\tabcolsep}{2pt} 

\begin{tabularx}{\linewidth}{l Y Y Y Y Y}
\toprule
\multirow{2}{*}{Scenario} & \multicolumn{5}{c}{\makecell{Normalized Top 10\% Mean RTL \\ at Penetration Rate (\%)}} \\
\cmidrule(lr){2-6}
& 25\% & 50\% & 75\% & 90\% & 100\% \\
\midrule
Changchun (Veh-Veh)   & 91.85 & 81.78 & 63.73 & 42.24 & 1.24 \\
Tianjin (Veh-Veh)     & 85.49 & 64.57 & 36.83 & 22.40 & 0.00 \\
Xi'an (Veh-Veh)       & 87.35 & 75.19 & 57.21 & 32.30 & 0.37 \\
Chongqing (Veh-Veh)   & 88.03 & 71.70 & 43.26 & 25.43 & 0.00 \\
Changchun (Veh-VRU)   & 91.61 & 81.22 & 62.64 & 42.02 & 0.02 \\
Tianjin (Veh-VRU)     & 90.68 & 74.73 & 45.86 & 20.25 & 0.00 \\
Xi'an (Veh-VRU)       & 89.16 & 78.35 & 52.49 & 37.96 & 0.00 \\
Chongqing (Veh-VRU)   & 87.90 & 77.72 & 50.52 & 29.34 & 0.00 \\
\midrule
Merging Lane (Veh-Veh)     & 87.83 & 71.74 & 51.32 & 34.40 & 24.65 \\
Ordinary Road (Veh-Veh)    & 89.16 & 74.99 & 57.73 & 30.58 & 5.00 \\
Multi-fork Road (Veh-Veh)  & 91.39 & 66.59 & 35.29 & 16.01 & 0.00 \\
T-junction (Veh-Veh)       & 93.15 & 81.78 & 55.11 & 37.45 & 0.00 \\
\bottomrule
\multicolumn{6}{l}{
  \begin{minipage}{\linewidth}
    \vspace{2pt}
    \scriptsize
    \textit{Note:} Values indicate the ratio of the Top 10\% Mean RTL at the given penetration rate to the corresponding baseline Top 10\% Mean RTL (0\% penetration).
  \end{minipage}
}
\end{tabularx}
\end{table}

Figure \ref{fig:3x4_grid} presents the RTL results for all scenarios using Complementary Cumulative Distribution Functions (CCDFs). To generate these plots, the RTL values for every Road User in each configuration were sorted in ascending order to compute the CCDF. The resulting curves, each corresponding to a different penetration rate, are plotted with RTL on the x-axis and the CCDF value on a logarithmic y-axis.

As illustrated in Figure \ref{fig:3x4_grid}, a consistent trend of risk reduction is observed across all scenarios for both Veh-Veh and Veh-VRU interactions as the penetration rate increases. However, this reduction is non-linear. The risk mitigation at lower penetration rates is modest; the safety benefit gained from the first 50\% of adoption is significantly smaller than that gained from the final 50\%. To provide a precise quantitative evaluation, Table \ref{tab:penetration} reports the Normalized Top $10\%$ Mean RTL across varying penetration rates. These numerical results substantiate the non-linear relationship observed in the CCDF plots. For instance, in the Changchun (Veh-Veh) scenario, the risk remains at 42.24\% of the baseline even at a 90\% penetration rate. A precipitous decline to 1.24\% is achieved only upon reaching 100\% adoption. These results characterize a non-linear risk-benefit profile, highlighting that the most substantial safety gains are realized only at very high levels of V2X adoption

At 100\% penetration, RTL values in the vast majority of scenarios are reduced to zero, indicating that V2X communication can nearly eliminate global risk. However, a critical exception exists. In certain Waymo scenarios, such as merging lanes and real-world environments with complex geometries, isolated vehicles can remain in the perception blind zones of all other participants due to irregular road alignments or persistent visual occlusions. As shown in Figure \ref{subfig:i} and \ref{subfig:j}, these globally unobserved agents result in non-zero residual risk even with full communication coverage, as V2X cannot share information that no participant possesses.

Moreover, relative risk levels at identical penetration rates are strongly correlated with traffic density. The Changchun scenario, characterized by high traffic density and frequent occlusions, requires a penetration rate exceeding 90–95\% to achieve significant risk reduction. This result indicates a general trend: denser traffic scenarios demand a higher penetration threshold for effective risk mitigation and exhibit a more delayed realization of safety benefits.

\subsection{Impact of Communication Strategies in Mixed Traffic}

During the phased, real-world deployment of V2X technology, traffic environments will not be homogeneous systems composed exclusively of autonomous vehicles. Instead, they will consist of mixed traffic, integrating human-driven (non-connected) vehicles with connected vehicles. To realistically model this transitional phase, we establish a heterogeneous configuration. In our evaluation, non-connected vehicles are modeled with a limited perception range, specifically a 120° forward-facing FoV. This constraint mirrors human drivers and basic forward-facing sensors. In contrast, connected vehicles are modeled with a full 360° circular FoV, representing the comprehensive sensing capabilities of advanced autonomous systems. This explicit perceptual disparity is crucial for assessing the safety benefits derived from collaborative perception as connectivity adoption increases.

\begin{figure}[t]
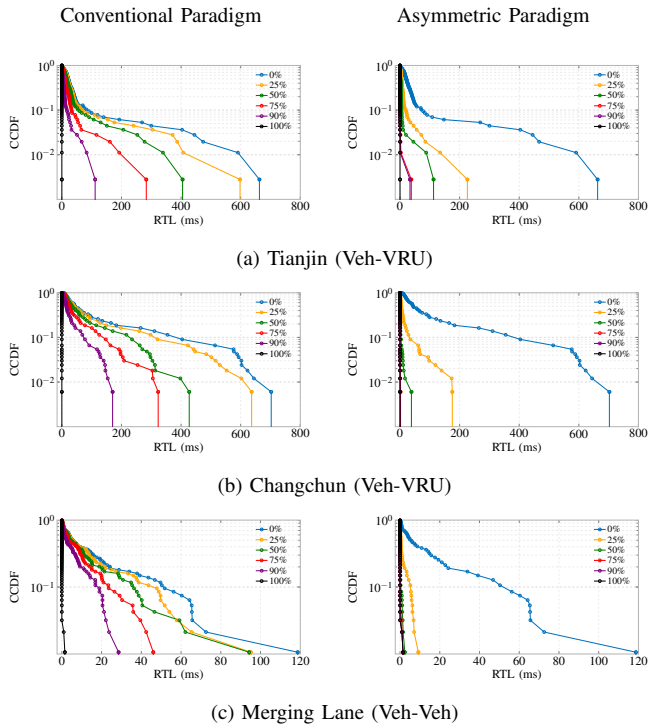

    \centering
    \vspace{0pt}
    \begin{tabular}{{c @{\hspace{1.8cm}} c}}
        \footnotesize{Conventional Paradigm} & \footnotesize{Asymmetric Paradigm} \\
    \end{tabular}
    \subfloat[Tianjin (Veh-VRU)]{%
        \begin{tabular}{cc}
            \resizebox{0.46\linewidth}{!}{%
                \input{tikztex/tianjin_veh-vru_dt500ms_dualFOV_wMTL-CCDF}
            }
            &
            \resizebox{0.46\linewidth}{!}{%
                \input{tikztex/tianjin_veh-vrunew_wMTL-CCDF}
            }
        \end{tabular}
        \label{fig:row_tianjin}
    }
    
    \vspace{-0.5em}
    
    \subfloat[Changchun (Veh-VRU)]{%
        \begin{tabular}{cc}
            \resizebox{0.46\linewidth}{!}{%
                \input{tikztex/changchun_veh-vru_dt500ms_dualFOV_wMTL-CCDF}
            }
            &
            \resizebox{0.46\linewidth}{!}{%
                \input{tikztex/changchun_veh-vru_dt500ms_dualFOV_wMTLnew-CCDF}
            }
        \end{tabular}
        \label{fig:row_changchun}
    }
    
    \vspace{-0.5em}

    \subfloat[Merging Lane (Veh-Veh)]{%
        \begin{tabular}{cc}
            \resizebox{0.46\linewidth}{!}{%
                \input{tikztex/huiru_traditional}
            }
            &
            \resizebox{0.46\linewidth}{!}{%
                \input{tikztex/newhuiru_veh-veh_dt500ms_dualFOV_wMTL-CCDF}
            }
        \end{tabular}
        \label{fig:row_merging} %
    }
    
    \caption{Comparison of risk mitigation benefits between conventional and asymmetric communication paradigms. \textit{Left column} evaluates the conventional communication paradigm, which relies on symmetric, bi-directional communication between connected vehicles; it thus requires high penetration rates for safety benefits. \textit{Right column} assesses the asymmetric V2V paradigm, where communication is not strictly symmetric. Non-connected vehicles can still receive perception data and warnings from connected vehicles. The asymmetric paradigm achieves superior safety benefits at lower penetration rates, with 25\% penetration outperforming 75-90\% in the conventional model.}
    \label{fig:combined_all}
\end{figure}

We first evaluate the conventional communication paradigm. In this model, communication is bidirectional but exclusive: connected vehicles can both transmit and receive perception data, but only with other connected vehicles. Human-driven, non-connected vehicles are perceptually and communicatively isolated, unable to either send or receive information. The safety performance of this conventional model, under varying penetration rates, is presented in Figure \ref{fig:combined_all} (left column). The results focus on three high-risk scenarios: Veh-VRU interactions in Tianjin and Changchun, and Veh-Veh interactions on the merging lane. A primary finding is that the safety benefit is highly non-linear and strictly dependent on a high penetration rate. The enhanced 360° FoV of connected vehicles provides a clear perceptual advantage over the 120° FoV of non-connected vehicles; however, this advantage is poorly leveraged at low adoption levels. As shown in the data, the risk reduction from the first 25\% of adoption is significantly smaller than that from the final 25\% increment (i.e., from 75\% to 100\%). This diminishing return profile confirms that the conventional V2V paradigm relies on a dense network effect. This high dependency creates a substantial economic and logistical barrier to initial deployment, as significant safety benefits are realized only after a majority of vehicles on the road are equipped with expensive bidirectional communication hardware.

To mitigate the prohibitive costs and slow adoption cycle of the conventional model, we introduce and evaluate our proposed asymmetric communication paradigm. This framework fundamentally redefines the flow of information in a mixed-traffic environment. The core concept is that while non-connected vehicles cannot transmit data, they can receive data broadcast by connected vehicles. In this model, connected vehicles act as mobile, high-fidelity perception hubs, continuously broadcasting their 360° environmental awareness. Non-connected vehicles, acting as "receive-only" nodes, can ingest this critical data. 

To illustrate, in a mixed-traffic urban scenario, conventional connected Robotaxis only share data among themselves. Our asymmetric paradigm differs: a Robotaxi broadcasts its rich perception data—such as an occluded pedestrian or a vehicle in a blind spot. A nearby non-connected, human-driven vehicle receives this broadcast. This vital information is then relayed to the human driver via a Human-Machine Interface, extending the Robotaxi's superior perception, bridging the perception gap, and mitigating immediate risks.

The transformative impact of this asymmetric approach is quantified in Figure \ref{fig:combined_all} (right column). A direct comparison with the conventional model reveals a dramatic improvement in safety outcomes, particularly at low penetration rates. The results are striking: a mere 25\% penetration rate under the asymmetric paradigm achieves safety benefits that surpass even a 75\% penetration rate in the conventional scheme across all three high-risk scenarios. This indicates an immediate and substantial return on investment. The benefit is particularly pronounced in the merging lane scenario, where the safety level at 25\% asymmetric penetration exceeds that of a 90\% conventional adoption rate. As the asymmetric penetration rate increases to 50\%, the risk reduction becomes overwhelming. In the merging lane scenario, the safety benefit effectively reaches saturation, performing comparably to a 100\% penetration level. In the Tianjin Veh-VRU scenario, the RTL values for over 90\% of VRUs are reduced to zero at this 50\% adoption level. By 75\% penetration, the Changchun scenario also approaches saturation, and 99\% of VRU risk in the Tianjin scenario is eliminated.

Furthermore, a comprehensive sensitivity analysis is conducted to verify the stability of the observed safety trends across varying parameter configurations. Table \ref{tab:param_sensitivity_analysis} summarizes the Top 10\% Mean RTL for both communication paradigms under multiple perturbations of risk coefficients, prediction horizons, and safety margins. Although absolute risk values fluctuate with the chosen operational thresholds, the relative performance rankings between the two paradigms remain invariant. Notably, the asymmetric paradigm at 25\% penetration consistently exhibits superior risk mitigation compared to the conventional paradigm at 75\% penetration across all tested variations. This stability confirms that the safety benefits are a structural property of the asymmetric information flow architecture, independent of specific parameter tuning.

The findings from this comparative analysis are clear. The conventional V2V paradigm, by restricting information flow to a closed network of connected vehicles, necessitates a prohibitively high penetration rate to manifest meaningful safety benefits. This presents a prohibitive deployment barrier, as the substantial initial costs are not justified by immediate benefits, which only manifest at high penetration rates. Our proposed asymmetric communication paradigm breaks this dependency. By enabling even non-connected vehicles to receive critical safety broadcasts, it achieves superior safety outcomes at fractional adoption levels. This model offers a highly cost-effective and practical implementation pathway. It significantly reduces the prohibitive initial investment in hardware—such as the requirement for on-board LiDAR and bidirectional communication equipment on every vehicle—by leveraging the broadcasts of a smaller subset of advanced connected vehicles. This proves the strong economic feasibility of our approach, accelerating the timeline for achieving comprehensive traffic safety.

\begin{table}[t]
\centering
\caption{Parameter Sensitivity Analysis: RTL Comparison Between Paradigms}
\label{tab:param_sensitivity_analysis}

\footnotesize
\renewcommand{\arraystretch}{1.1}
\setlength{\tabcolsep}{2pt} 

\begin{tabularx}{\linewidth}{l Y Y Y Y Y Y}
\toprule
\multirow{2}{*}{Parameter} & \multicolumn{6}{c}{Top 10\% Mean RTL at Penetration Rate (ms)} \\
\cmidrule(lr){2-7}
& 0\% & 25\% & 50\% & 75\% & 90\% & 100\% \\
\midrule
\multicolumn{7}{l}{\textit{\textbf{Conventional Paradigm}}} \\ 
Baseline                             & 124.33 & 103.77 & 59.48 & 42.60 & 15.78 & 0.00 \\
K1 ($k_{\mathrm{overlap}} \downarrow$)        & 76.67 & 63.18 & 41.04 & 27.15 & 17.15 & 0.00 \\
K2 ($k_{\mathrm{overlap}} \uparrow$)          & 203.47 & 162.32 & 102.86 & 69.51 & 36.35 & 0.00 \\
K3 ($k_{\mathrm{no\mbox{-}overlap}} \downarrow$) & 112.29 & 81.71 & 60.48 & 29.22 & 27.10 & 0.00 \\
K4 ($k_{\mathrm{no\mbox{-}overlap}} \uparrow$)   & 168.29 & 142.27 & 105.73 & 65.88 & 30.48 & 0.00 \\
K5 ($k_{\mathrm{static}} \uparrow$)         & 131.81 & 106.40 & 82.41 & 48.50 & 24.48 & 0.00 \\
$M_s$ (margin $\downarrow$)            & 65.03 & 52.78 & 42.45 & 23.90 & 10.69 & 0.00 \\
$M_l$ (margin $\uparrow$)               & 253.39 & 212.89 & 170.22 & 92.30 & 50.68 & 0.00 \\
$D_s$ (horizon $\downarrow$)         & 83.58 & 67.92 & 53.01 & 31.31 & 14.36 & 0.00 \\
$D_l$ (horizon $\uparrow$)           & 135.62 & 110.60 & 79.31 & 45.71 & 20.24 & 0.00 \\

\midrule

\multicolumn{7}{l}{\textit{\textbf{Asymmetric Paradigm}}} \\ 
Baseline                             & 124.33 & 18.07 & 4.42 & 0.26 & 0.02 & 0.00 \\
K1 ($k_{\mathrm{overlap}} \downarrow$)        & 76.67 & 13.93 & 2.51 & 0.07 & 0.03 & 0.00 \\
K2 ($k_{\mathrm{overlap}} \uparrow$)          & 203.47 & 38.30 & 12.58 & 0.20 & 0.08 & 0.00 \\
K3 ($k_{\mathrm{no\mbox{-}overlap}} \downarrow$) & 112.29 & 3.75 & 0.67 & 0.13 & 0.03 & 0.00 \\
K4 ($k_{\mathrm{no\mbox{-}overlap}} \uparrow$)   & 168.29 & 23.49 & 6.70 & 0.86 & 0.01 & 0.00 \\
K5 ($k_{\mathrm{static}} \uparrow$)         & 131.81 & 26.48 & 4.33 & 1.25 & 0.17 & 0.00 \\
$M_s$ (margin $\downarrow$)             & 65.03 & 10.93 & 3.53 & 1.21 & 0.00 & 0.00 \\
$M_l$ (margin $\uparrow$)               & 253.39 & 39.64 & 9.40 & 0.11 & 0.01 & 0.00 \\
$D_s$ (horizon $\downarrow$)         & 83.58 & 17.21 & 1.25 & 0.13 & 0.03 & 0.00 \\
$D_l$ (horizon $\uparrow$)           & 135.62 & 17.22 & 4.07 & 0.75 & 0.04 & 0.00 \\

\bottomrule
\multicolumn{7}{l}{
  \begin{minipage}{\linewidth}
    \vspace{2pt}
    \scriptsize
    \textit{Note:} Baseline parameters are configured as $k_{\mathrm{overlap}}$: 3.0, 1.0; $k_{\mathrm{no\mbox{-}overlap}}$: 0.4, 0.2; $k_{\mathrm{static}}$: 0.4, 0.2;  $\text{safety margin }M$: 0.3m; and $\text{prediction horizon }$: 600ms. 
    K1--K5 denote coefficient variations: 
    K1 ($k_{\mathrm{overlap}} \!\to\!$ 0.8, 0.6); 
    K2 ($k_{\mathrm{overlap}} \!\to\!$ 5.0, 3.0); 
    K3 ($k_{\mathrm{no\mbox{-}overlap}} \!\to\!$ 0.1, 0.08); 
    K4 ($k_{\mathrm{no\mbox{-}overlap}} \!\to\!$ 0.8, 0.6); 
    K5 ($k_{\mathrm{static}} \!\to\!$ 0.08, 0.05).
    $M_s/M_l$: $\text{safety margin}\!\to\!$ 0.1/0.5\,m; $D_s/D_l$: $\text{prediction horizon} \!\to\!$ 100/1000\,ms.
     Experiments were conducted under the Tianjin Veh-Veh scenario with a 120° FoV assumption.
  \end{minipage}
}
\end{tabularx}
\end{table}

 \section{Conclusions}
This paper introduced a framework to quantify occlusion-induced collision risk, bridging the gap between V2X perception and quantifiable safety gains. The framework is centered on our novel RTL metric. Validation results confirmed that RTL delivers a holistic risk assessment, adeptly capturing both high-intensity, transient threats and prolonged, low-intensity exposure. Crucially, this metric demonstrates a strong correlation with real-world traffic hazards. Furthermore, we proposed a novel asymmetric communication paradigm allowing non-connected vehicles to receive safety broadcasts. This approach surmounts key V2X deployment barriers, achieving superior safety benefits at lower penetration rates and thus providing a more cost-effective, viable roadmap. 

Future research will incorporate realistic communication models to assess the impact of latency and packet loss on risk quantification. Additionally, the practical implementation of the asymmetric paradigm requires investigating Human-Machine Interface designs for effectively relaying safety broadcasts to human drivers. Although these implementation challenges remain, the RTL metric and the proposed roadmap provide a robust foundation and a cost-effective pathway for accelerating V2X safety benefits in mixed-traffic environments.

\bibliographystyle{IEEEtran}
\bibliography{reference}

@INPROCEEDINGS{Wolff2024,
  author={Wolff, Vincent Albert and Xhoxhi, Edmir},
  booktitle={2024 IEEE Intelligent Vehicles Symposium (IV)}, 
  title={Mitigating Vulnerable Road Users Occlusion Risk Via Collective Perception: An Empirical Analysis}, 
  year={2024},
  volume={},
  number={},
  pages={124-128}}

@article{gilroy2019overcoming,
  title={Overcoming occlusion in the automotive environment—A review},
  author={Gilroy, Shane and Jones, Edward and Glavin, Martin},
  journal={IEEE Transactions on Intelligent Transportation Systems},
  volume={22},
  number={1},
  pages={23--35},
  year={2019},
  publisher={IEEE}
}

@article{kumar2023surround,
  title={Surround-view fisheye camera perception for automated driving: Overview, survey \& challenges},
  author={Kumar, Varun Ravi and Eising, Ciar{\'a}n and Witt, Christian and Yogamani, Senthil Kumar},
  journal={IEEE Transactions on Intelligent Transportation Systems},
  volume={24},
  number={4},
  pages={3638--3659},
  year={2023},
  publisher={IEEE}
}

@article{caillot2022survey,
  title={Survey on cooperative perception in an automotive context},
  author={Caillot, Antoine and Ouerghi, Safa and Vasseur, Pascal and Boutteau, R{\'e}mi and Dupuis, Yohan},
  journal={IEEE Transactions on Intelligent Transportation Systems},
  volume={23},
  number={9},
  pages={14204--14223},
  year={2022},
  publisher={IEEE}
}

@inproceedings{xu2022drone,
  title={SIND: A drone dataset at signalized intersection in China},
  author={Xu, Yanchao and Shao, Wenbo and Li, Jun and Yang, Kai and Wang, Weida and Huang, Hua and Lv, Chen and Wang, Hong},
  booktitle={2022 IEEE 25th International Conference on Intelligent Transportation Systems (ITSC)},
  pages={2471--2478},
  year={2022},
  organization={IEEE}
}

@inproceedings{xu2023v2v4real,
  title={V2v4real: A real-world large-scale dataset for vehicle-to-vehicle cooperative perception},
  author={Xu, Runsheng and Xia, Xin and Li, Jinlong and Li, Hanzhao and Zhang, Shuo and Tu, Zhengzhong and Meng, Zonglin and Xiang, Hao and Dong, Xiaoyu and Song, Rui and others},
  booktitle={Proceedings of the IEEE/CVF conference on computer vision and pattern recognition},
  pages={13712--13722},
  year={2023}
}

@article{yu2019occlusion,
  title={Occlusion-aware risk assessment for autonomous driving in urban environments},
  author={Yu, Ming-Yuan and Vasudevan, Ram and Johnson-Roberson, Matthew},
  journal={IEEE Robotics and Automation Letters},
  volume={4},
  number={2},
  pages={2235--2241},
  year={2019},
  publisher={IEEE}
}

@inproceedings{song2025traf,
  title={Traf-align: Trajectory-aware feature alignment for asynchronous multi-agent perception},
  author={Song, Zhiying and Yang, Lei and Wen, Fuxi and Li, Jun},
  booktitle={Proceedings of the Computer Vision and Pattern Recognition Conference},
  pages={12048--12057},
  year={2025}
}

@ARTICLE{10919014,
  author={Yang, Lei and Tang, Tao and Li, Jun and Yuan, Kun and Wu, Kai and Chen, Peng and Wang, Li and Huang, Yi and Li, Lei and Zhang, Xinyu and Yu, Kaicheng},
  journal={IEEE Transactions on Pattern Analysis and Machine Intelligence}, 
  title={BEVHeight++: Toward Robust Visual Centric 3D Object Detection}, 
  year={2025},
  volume={47},
  number={6},
  pages={5094-5111},
  doi={10.1109/TPAMI.2025.3549711}}

@inproceedings{xu2022v2x,
  title={V2x-vit: Vehicle-to-everything cooperative perception with vision transformer},
  author={Xu, Runsheng and Xiang, Hao and Tu, Zhengzhong and Xia, Xin and Yang, Ming-Hsuan and Ma, Jiaqi},
  booktitle={European conference on computer vision},
  pages={107--124},
  year={2022},
  organization={Springer}
}

@ARTICLE{11096563,
  author={Xie, Tenghui and Song, Zhiying and Wen, Fuxi and Li, Jun and Liu, Guangzhao and Zhao, Zijian},
  journal={IEEE Robotics and Automation Letters}, 
  title={TruckV2X: A Truck-Centered Perception Dataset}, 
  year={2025},
  volume={10},
  number={9},
  pages={9312-9319},
  doi={10.1109/LRA.2025.3592884}}

@inproceedings{10.1145/3498361.3538925,
author = {Qiu, Hang and Huang, Po-Han and Asavisanu, Namo and Liu, Xiaochen and Psounis, Konstantinos and Govindan, Ramesh},
title = {AutoCast: scalable infrastructure-less cooperative perception for distributed collaborative driving},
year = {2022},
isbn = {9781450391856},
publisher = {Association for Computing Machinery},
address = {New York, NY, USA},
url = {https://doi.org/10.1145/3498361.3538925},
doi = {10.1145/3498361.3538925},
booktitle = {Proceedings of the 20th Annual International Conference on Mobile Systems, Applications and Services},
pages = {128–141},
numpages = {14},
location = {Portland, Oregon},
series = {MobiSys '22}
}

@inproceedings{cui2022coopernaut,
  title={Coopernaut: End-to-end driving with cooperative perception for networked vehicles},
  author={Cui, Jiaxun and Qiu, Hang and Chen, Dian and Stone, Peter and Zhu, Yuke},
  booktitle={Proceedings of the IEEE/CVF Conference on Computer Vision and Pattern Recognition},
  pages={17252--17262},
  year={2022}
}

@article{deng2019cooperative,
  title={Cooperative collision avoidance for overtaking maneuvers in cellular V2X-based autonomous driving},
  author={Deng, Ruoqi and Di, Boya and Song, Lingyang},
  journal={IEEE Transactions on Vehicular Technology},
  volume={68},
  number={5},
  pages={4434--4446},
  year={2019},
  publisher={IEEE}
}

@inproceedings{wang2024deepaccident,
  title={Deepaccident: A motion and accident prediction benchmark for v2x autonomous driving},
  author={Wang, Tianqi and Kim, Sukmin and Wenxuan, Ji and Xie, Enze and Ge, Chongjian and Chen, Junsong and Li, Zhenguo and Luo, Ping},
  booktitle={Proceedings of the AAAI Conference on Artificial Intelligence},
  volume={38},
  number={6},
  pages={5599--5606},
  year={2024}
}

@inproceedings{sun2020scalability,
  title={Scalability in perception for autonomous driving: Waymo open dataset},
  author={Sun, Pei and Kretzschmar, Henrik and Dotiwalla, Xerxes and Chouard, Aurelien and Patnaik, Vijaysai and Tsui, Paul and Guo, James and Zhou, Yin and Chai, Yuning and Caine, Benjamin and others},
  booktitle={Proceedings of the IEEE/CVF conference on computer vision and pattern recognition},
  pages={2446--2454},
  year={2020}
}

@article{park2023occlusion,
  title={Occlusion-aware risk assessment and driving strategy for autonomous vehicles using simplified reachability quantification},
  author={Park, Hyunwoo and Choi, Jongseo and Chin, Hyuntai and Lee, Sang-Hyun and Baek, Doosan},
  journal={IEEE Robotics and Automation Letters},
  volume={8},
  number={12},
  pages={8486--8493},
  year={2023},
  publisher={IEEE}
}

@article{chen2025occlusion,
  title={Occlusion-aware trajectory planning with quantified risk constraint for deadlock mitigation in autonomous driving},
  author={Chen, Zhan and Liu, Weidong and Xiong, Lu and Yu, Zhuoping and Tang, Chen},
  journal={IEEE Transactions on Intelligent Transportation Systems},
  year={2025},
  publisher={IEEE}
}

@article{koschi2020set,
  title={Set-based prediction of traffic participants considering occlusions and traffic rules},
  author={Koschi, Markus and Althoff, Matthias},
  journal={IEEE Transactions on Intelligent Vehicles},
  volume={6},
  number={2},
  pages={249--265},
  year={2020},
  publisher={IEEE}
}

@article{mcgill2019probabilistic,
  title={Probabilistic risk metrics for navigating occluded intersections},
  author={McGill, Stephen G and Rosman, Guy and Ort, Teddy and Pierson, Alyssa and Gilitschenski, Igor and Araki, Brandon and Fletcher, Luke and Karaman, Sertac and Rus, Daniela and Leonard, John J},
  journal={IEEE Robotics and Automation Letters},
  volume={4},
  number={4},
  pages={4322--4329},
  year={2019},
  publisher={IEEE}
}

@article{zhang2023occlusion,
  title={Occlusion-aware planning for autonomous driving with vehicle-to-everything communication},
  author={Zhang, Chi and Steinhauser, Florian and Hinz, Gereon and Knoll, Alois},
  journal={IEEE Transactions on Intelligent Vehicles},
  volume={9},
  number={1},
  pages={1229--1242},
  year={2023},
  publisher={IEEE}
}

@article{pan2025lateral,
  title={Lateral Displacement and Distance of Vehicles in Freeway Overtaking Scenario Based on Naturalistic Driving Data.},
  author={Pan, Cunshu and Zhang, Yuhao and Zhang, Heshan and Xu, Jin},
  journal={Applied Sciences (2076-3417)},
  volume={15},
  number={5},
  year={2025}
}

@article{laberge2009occupant,
  title={Occupant injury severity from lateral collisions: a literature review},
  author={Laberge-Nadeau, Claire and Bellavance, Fran{\c{c}}ois and Messier, St{\'e}phane and V{\'e}zina, Lyne and Pichette, Fernand},
  journal={Journal of safety research},
  volume={40},
  number={6},
  pages={427--435},
  year={2009},
  publisher={Elsevier}
}

@article{garcia2021tutorial,
  title={A tutorial on 5G NR V2X communications},
  author={Garcia, Mario H Casta{\~n}eda and Molina-Galan, Alejandro and Boban, Mate and Gozalvez, Javier and Coll-Perales, Baldomero and {\c{S}}ahin, Taylan and Kousaridas, Apostolos},
  journal={IEEE Communications Surveys \& Tutorials},
  volume={23},
  number={3},
  pages={1972--2026},
  year={2021},
  publisher={IEEE}
}

@article{manzinger2020using,
  title={Using reachable sets for trajectory planning of automated vehicles},
  author={Manzinger, Stefanie and Pek, Christian and Althoff, Matthias},
  journal={IEEE Transactions on Intelligent Vehicles},
  volume={6},
  number={2},
  pages={232--248},
  year={2020},
  publisher={IEEE}
}

@article{green2000long,
  title={" How long does it take to stop?" Methodological analysis of driver perception-brake times},
  author={Green, Marc},
  journal={Transportation human factors},
  volume={2},
  number={3},
  pages={195--216},
  year={2000},
  publisher={Taylor \& Francis}
}

@article{shalev2017formal,
  title={On a formal model of safe and scalable self-driving cars},
  author={Shalev-Shwartz, Shai and Shammah, Shaked and Shashua, Amnon},
  journal={arXiv preprint arXiv:1708.06374},
  year={2017}
}

@article{johansson1971drivers,
  title={Drivers' brake reaction times},
  author={Johansson, Gunnar and Rumar, K{\aa}re},
  journal={Human factors},
  volume={13},
  number={1},
  pages={23--27},
  year={1971},
  publisher={SAGE Publications Sage CA: Los Angeles, CA}
}


 





\end{document}